\documentclass[11pt]{article}

\usepackage[a4paper, margin=1in]{geometry}
\usepackage[T1]{fontenc}
\usepackage[utf8]{inputenc}
\usepackage{lmodern}
\usepackage{microtype}
\usepackage{textcomp}

\usepackage{amsmath, amssymb, amsfonts}

\usepackage{graphicx}
\usepackage{booktabs}
\usepackage{tabularx}
\usepackage{array}
\usepackage{float}
\usepackage{placeins}

\usepackage{xcolor}
\definecolor{linkcolor}{rgb}{0.55,0.14,0.20}
\definecolor{codebg}{rgb}{0.985,0.975,0.945}

\usepackage[colorlinks=true,
            linkcolor=linkcolor,
            citecolor=linkcolor,
            urlcolor=linkcolor]{hyperref}
\usepackage{url}

\usepackage{fancyhdr}
\usepackage{titlesec}
\titleformat{\section}{\large\bfseries\sffamily}{\thesection}{0.5em}{}
\titleformat{\subsection}{\normalsize\bfseries\sffamily}{\thesubsection}{0.5em}{}
\titleformat{\subsubsection}{\normalsize\bfseries\sffamily}{\thesubsubsection}{0.5em}{}
\titlespacing*{\section}{0pt}{8pt}{4pt}
\titlespacing*{\subsection}{0pt}{6pt}{2pt}
\titlespacing*{\subsubsection}{0pt}{6pt}{2pt}

\usepackage{caption}
\newcommand{\taxo}[1]{\texttt{#1}}

\title{\bfseries Paper Pilot: A Human-in-the-Loop Expert System for\\
       Evidence-Traceable Scientific Manuscript Generation\\
       in Applied Sciences}

\author{
  Nidhi Jha\thanks{Corresponding author: \href{mailto:nidhi.jha@uah.edu}{nidhi.jha@uah.edu}.}
  \quad and \quad
  Siddharth Chaudhary\thanks{Corresponding author: \href{mailto:siddharth.chaudhary@uah.edu}{siddharth.chaudhary@uah.edu}.}
  \quad and \quad
  Ajinkya Kulkarni\thanks{\href{mailto:ajinkya.kulkarni@uah.edu}{ajinkya.kulkarni@uah.edu}.}\\
  University of Alabama in Huntsville
}

\date{}

\begin{document}
\maketitle
\thispagestyle{fancy}

\begin{abstract}
\noindent
Large language model-based agents are increasingly being integrated
into scientific workflows for literature analysis, hypothesis
generation, code development, result interpretation, manuscript
drafting, and automated review. Existing systems demonstrate progress
toward autonomous scientific discovery and automated manuscript
generation; however, they do not fully resolve the governance problem
introduced when ideas, methods, results, and claims propagate through
AI-assisted workflows without mandatory human approval or
artifact-level traceability. This paper proposes \emph{Paper Pilot},
a human-in-the-loop expert system for evidence-traceable scientific
manuscript generation in applied sciences. The system adapts the
Collaborative Agent Reasoning Engineering (CARE) methodology to
manuscript development by introducing manuscript-owner approval
gates, explicit no-pass criteria, claim classification, audit
logging, advisory LLM review, and evidence-locked revision control.
Paper Pilot is implemented through ChatGPT, Gemini, Claude, or any
institutional LLM environment, with the system prompt intended for
open release through GitHub. The framework defines approval gates
across the idea-to-claim pipeline, including scope approval,
literature approval, method approval, evidence approval, claim
approval, reviewer override, section approval, and final release. It
also distinguishes literature-grounded claims from artifact-grounded
claims, requiring reported numbers, plots, tables, and
interpretations to remain traceable to approved evidence. A workflow
demonstration illustrates how the system supports section-wise
manuscript drafting, gap identification, claim classification,
revision control, and manuscript-owner oversight. As a first
empirical validation, we evaluate the citation-grounding layer with a
controlled, mechanically scored benchmark (two commercial LLMs, real
arXiv papers, arXiv-API verification, no LLM judge): under coverage
pressure ungated drafters fabricated up to 25\% of their citations and
never flagged an evidence gap, whereas the same models under Paper
Pilot's evidence-locked rules produced zero fabricated citations and
surfaced the planted gaps as explicit placeholders. Preliminary
results for result grounding, evidence-locked revision, and
adversarial robustness point the same way, and their full
evaluation---on real datasets and with the complete released
prompt---is left to future work. The proposed
framework positions LLM-assisted writing as a controlled human-AI
decision-support process rather than a fully autonomous authorship
pipeline.

\end{abstract}

\section{Introduction}
\label{sec:intro}

Large language models (LLMs) are increasingly being embedded into
agentic systems that support scientific research activities
traditionally performed by human researchers, including hypothesis
generation, literature analysis, experimental planning, code
development, result interpretation, manuscript drafting, and
automated review. This development has shifted AI-assisted research
from isolated task automation toward increasingly integrated
scientific workflows in which multiple agents may coordinate across
stages of knowledge production and scholarly communication. Recent
systems illustrate this transition. The AI Scientist was proposed as
an end-to-end framework capable of generating research ideas, writing
code, executing experiments, visualizing results, drafting complete
scientific papers, and applying a simulated review process for
evaluation~\cite{lu2024aiscientist}. The AI Scientist-v2 further
advanced this direction through agentic tree search, autonomous
experimental design, manuscript generation, and workshop-level paper
submission~\cite{yamada2025aiscientistv2}. Similarly, InternAgent was
introduced as a closed-loop multi-agent framework for autonomous
scientific research from hypothesis generation to experimental
verification~\cite{internagent2025}, while Agent Laboratory
demonstrated the use of LLM agents as research assistants across the
research process~\cite{schmidgall2025agentlab}.

In parallel, specialized systems have emerged for AI-assisted
scientific writing and scholarly synthesis. PaperOrchestra addresses
the transformation of unconstrained pre-writing materials into
submission-ready LaTeX manuscripts through a multi-agent writing
framework and evaluates this capability using
PaperWritingBench~\cite{song2026paperorchestra}. ARISE introduces an
agentic, rubric-guided iterative survey engine that uses
citation-first retrieval, citation-keyed memory, specialized writing
agents, and reviewer agents to generate scholarly survey
papers~\cite{wang2025arise}. Related work has also emphasized
scientific agents as collaborative or ecosystem-level systems.
OmniScientist argues that scientific discovery should not be treated
only as a standalone search or optimization problem because
real-world science depends on collaborative mechanisms, contribution
attribution, peer review, and structured knowledge
infrastructures~\cite{shao2025omniscientist}. These studies
collectively indicate that LLM-based scientific agents are moving
toward broader participation in the research lifecycle, including
both knowledge production and manuscript production.

Despite this progress, a critical control problem remains
insufficiently addressed. Existing systems often include human
interaction, automated reviewer agents, or citation-grounded writing
mechanisms, but they do not define a standard, mandatory, end-to-end
set of human approval gates across the closed research loop from idea
generation to final manuscript claims. In autonomous or
semi-autonomous scientific workflows, unsupported decisions can
propagate across stages: a weak research idea may lead to an invalid
method; an invalid method may produce unreliable code or results;
unreliable results may be converted into manuscript claims; and
manuscript claims may then be polished into publication-ready
language without sufficient evidentiary support. This propagation
risk is especially important for expert and intelligent systems
because the purpose of such systems is not only to generate outputs,
but also to support reliable decision-making, controllable
implementation, and accountable management in practical domains.

The need for such control is consistent with the Collaborative Agent
Reasoning Engineering (CARE) methodology, which frames LLM-agent
development as a disciplined, stage-gated engineering process rather
than ad hoc prompt iteration. CARE specifies agent behavior,
grounding, tool orchestration, and verification through reusable
artifacts, and organizes development through a three-party workflow
involving subject-matter experts, developers, and LLM-based helper
agents~\cite{ramachandran2026care}. In CARE, helper agents convert
informal domain intent into structured, reviewable artifacts that are
approved by humans at defined gates. This artifact-driven and
approval-gated orientation provides a methodological foundation for
designing scientific writing agents whose behavior is specifiable,
testable, maintainable, and auditable over time.

Building on this methodology, this paper proposes \emph{Paper Pilot},
a human-in-the-loop expert system for evidence-traceable scientific
manuscript generation in applied sciences. Paper Pilot adapts the
CARE methodology to the scientific writing lifecycle by treating the
manuscript owner as the accountable human authority for approval,
override, and release decisions. The system is implemented through
general-purpose and domain-oriented LLM environments and its system
prompt is made openly available through GitHub. In this
implementation, LLM agents assist with section-wise drafting,
literature-grounding checks, claim classification, revision control,
and gap identification, while the manuscript owner retains authority
over research validity, evidentiary sufficiency, interpretation, and
final submission decisions.

Beyond this absence of mandatory approval gates, a second limitation
concerns the authority assigned to LLM-based
reviewers or judges. Several systems use reviewer agents or automated
evaluation mechanisms to assess generated ideas, papers, figures, or
outputs. The AI Scientist incorporates an automated reviewer intended
to approximate human paper evaluation~\cite{lu2024aiscientist}, while
CycleResearcher uses CycleReviewer as part of an iterative
research-and-review process~\cite{weng2024cycleresearcher}. Such
reviewer agents may improve scalability and provide rapid feedback,
but they also raise unresolved questions regarding when human
authority must override LLM judgment, how reviewer drift should be
detected over time, and how systematic bias in automated evaluation
should be audited. In Paper Pilot, LLM reviewers are therefore
treated as advisory components rather than final authorities. Their
judgments may inform manuscript revision, but approval of scientific
claims, methodological adequacy, evidence sufficiency, and release
readiness remains assigned to the manuscript owner.

A third limitation concerns evidence-to-claim traceability.
Citation-grounded systems such as ARISE provide mechanisms for
linking scholarly statements to retrieved literature through
citation-keyed memory and evidence-aware revision
processes~\cite{wang2025arise}. However, scientific manuscripts also
contain computational and analytical claims that cannot be fully
validated through bibliographic citation alone. Reported numbers,
plots, tables, benchmark comparisons, ablation findings, and
statistical conclusions require traceability to run-level artifacts,
including code versions, data hashes, random seeds, environment
specifications, execution logs, configuration files, intermediate
outputs, and generated figures. Without such traceability, a writing
agent may introduce unsupported metrics, alter comparative
statements, or summarize experimental results in ways that are
linguistically plausible but not reproducibly grounded.

Paper Pilot addresses these gaps by formalizing a human-in-the-loop
approval architecture for LLM-assisted manuscript development. The
framework defines eight manuscript-owner approval gates across the
idea-to-claim pipeline, including scope approval, literature
approval, method approval, evidence approval, claim approval,
reviewer override, section approval, and final manuscript release.
Each gate is associated with explicit no-pass criteria, audit-log
requirements, and conditions under which LLM-generated or
LLM-reviewed content must be revised, rejected, or escalated. The
framework further extends citation-grounded writing toward
artifact-grounded writing by requiring reported claims to be linked
to either literature evidence, user-provided results, or inspectable
computational artifacts.

The proposed framework is designed for the applied
intelligent-systems context targeted by \emph{Expert Systems with
Applications}. It conceptualizes scientific manuscript generation as
a decision-support workflow in which LLM agents perform drafting,
checking, retrieval, review, and revision-support tasks, while the
manuscript owner retains authority over scientific validity,
methodological adequacy, evidentiary sufficiency, and final claims.
This positioning is distinct from fully autonomous paper-generation
systems because the objective is not to remove human researchers from
the loop, but to define where and how human judgment must be inserted
into the loop to preserve accountability, reproducibility, and
scientific integrity.

The research questions addressed in this paper are as follows:

\medskip
\noindent\textbf{RQ1.} What mandatory human approval gates, no-pass
criteria, and audit-log requirements should govern the transition
from idea to manuscript claim in LLM-assisted applied-science
writing, and how can these be specified as a reusable expert-system
workflow?

\medskip
\noindent\textbf{RQ2.} How should the authority boundary between LLM
reviewer agents and human manuscript owners be defined in an
approval-gated writing workflow, and what categories of decisions
must remain exclusively with the human author?

\medskip
The main contributions of this paper are fivefold. First, it
introduces Paper Pilot as a CARE-informed, human-in-the-loop expert
system for LLM-assisted scientific manuscript generation in applied
sciences. Second, it defines a manuscript-owner approval-gate model
across the idea-to-claim pipeline. Third, it specifies no-pass
criteria and audit-log requirements for controlling the progression
of AI-generated content through drafting, review, and revision
stages. Fourth, it proposes an artifact-level evidence-to-claim
traceability mechanism and evidence-locked revision process for
preventing unsupported metrics, unverifiable comparisons, and
unapproved interpretive claims from entering the manuscript. Fifth,
it provides a first empirical validation of the citation-grounding
layer through a controlled, judge-free benchmark on real arXiv papers
(Section~\ref{sec:eval}), with preliminary results for result
grounding, evidence-locked revision, and adversarial robustness and a
released benchmark suite for their fuller future evaluation.

\section{Related Work}
\label{sec:related}

\subsection{LLM-based agents for autonomous scientific discovery}

LLM-based scientific agents have increasingly been proposed as
systems capable of performing research activities beyond isolated
text generation. Early agentic research systems generally focused on
bounded assistance tasks, such as literature review, ideation, code
generation, or analysis support. More recent systems have attempted
to integrate these capabilities into closed-loop scientific workflows
that include hypothesis generation, experimental execution,
manuscript preparation, and automated review. The AI Scientist
represents one of the clearest examples of this trajectory. It was
designed as a framework for fully automated scientific discovery in
machine learning, including idea generation, code writing, experiment
execution, visualization, paper writing, and simulated
review~\cite{lu2024aiscientist}. The AI Scientist-v2 further extends
this direction by using agentic tree search for autonomous scientific
discovery and manuscript generation~\cite{yamada2025aiscientistv2}.

Closed-loop scientific automation has also been explored through
broader multi-agent frameworks. InternAgent was proposed as a unified
closed-loop multi-agent framework for autonomous scientific research
across multiple scientific fields, connecting hypothesis generation
with verification through coordinated agents~\cite{internagent2025}.
OmniScientist expands the scope further by arguing that AI scientists
should be embedded within a co-evolving human-AI scientific ecosystem
rather than treated only as isolated optimization systems. Its design
emphasizes citation networks, collaboration protocols, peer review,
contribution attribution, and structured scientific knowledge
infrastructure~\cite{shao2025omniscientist}. Survey work on agentic
AI for scientific discovery similarly indicates that AI agents are
increasingly being used across literature review, hypothesis
generation, experiment planning, data analysis, and scientific
reasoning tasks across domains such as chemistry, biology, and
materials science~\cite{gridach2025agentic}.

Additional work has explored collaborative infrastructure and
co-scientist models for agentic research. \emph{Towards an AI
co-scientist} frames LLM-based systems as collaborative partners for
scientific reasoning and hypothesis generation rather than as isolated
automation tools~\cite{gottweis2025coscientist}. AgentRxiv proposes a
shared research infrastructure in which autonomous agents can upload,
retrieve, and build upon research reports, highlighting the need for
coordination and traceability in agentic research
ecosystems~\cite{schmidgall2025agentrxiv}.

These systems demonstrate the feasibility and growing ambition of
automated scientific workflows. However, their primary objective is
often to increase the autonomy, scope, or efficiency of AI-driven
discovery. Paper Pilot addresses a complementary problem: how to make
LLM-assisted scientific manuscript generation governable, auditable,
and evidence-traceable when human authority remains mandatory. Rather
than treating human oversight as a general interaction channel, Paper
Pilot formalizes manuscript-owner approval gates across the
idea-to-claim pipeline. This distinction is central for
applied-science settings, where research outputs may depend on
heterogeneous methods, domain-specific assumptions, and
context-sensitive evidence requirements.

\subsection{Automated scientific writing and scholarly synthesis
systems}

A second line of related work focuses more directly on scientific
manuscript generation and scholarly synthesis. PaperOrchestra
proposes a multi-agent framework for automated AI research paper
writing that transforms unconstrained pre-writing materials into
submission-ready LaTeX manuscripts. It also introduces
PaperWritingBench, a benchmark constructed from reverse-engineered
raw materials from top-tier AI conference papers, and reports
improvements over autonomous baselines in literature review and
manuscript quality~\cite{song2026paperorchestra}. ARISE focuses on
automated scholarly survey generation through an agentic
rubric-guided iterative survey engine. Its architecture includes
specialized LLM agents for topic expansion, citation curation,
literature summarization, manuscript drafting, and peer-review-based
evaluation, with citation-first retrieval and citation-keyed memory
supporting literature grounding~\cite{wang2025arise}.

Automated survey-generation systems also contribute to this
literature by improving scholarly synthesis at scale. AutoSurvey2,
for example, is positioned as an automated literature-survey system
intended to support researchers in constructing higher-level
literature reviews from scholarly sources~\cite{song2025autosurvey2}.
Such systems are relevant because they illustrate how LLM agents can
assist with synthesis, organization, and scholarly framing. However,
automated survey generation remains distinct from approval-gated
manuscript governance. A literature-survey system can support source
aggregation and synthesis, but it does not necessarily define
mandatory human gates for authorizing methods, results, claims, and
revisions.

These systems make important contributions to manuscript-generation
quality, literature synthesis, and iterative refinement. They also
show that multi-agent architectures can produce more structured
scientific writing than single-prompt drafting. Nevertheless,
automated writing quality and evidence governance are distinct
objectives. A manuscript may be coherent, well formatted, and
citation-rich while still containing computational claims that are
insufficiently linked to run artifacts or results supplied by the
manuscript owner. Citation grounding can reduce unsupported
literature claims, but it does not by itself ensure that reported
metrics, plots, ablation findings, experimental comparisons, or
analytical conclusions are reproducibly linked to code, data, seeds,
environments, logs, and outputs.

Paper Pilot therefore extends the scholarly-writing-agent literature
by introducing a result-grounding layer in addition to literature
grounding. Its core design requirement is that manuscript text should
not be revised independently of the evidence state. Reported claims
must remain linked to literature evidence, user-provided findings, or
computational artifacts. When a writing or reviewer agent introduces
a metric, comparison, or interpretation that lacks traceable support,
the revision must fail the relevant gate and return to the manuscript
owner for approval, correction, or rejection.

\subsection{LLM reviewers, automated judges, and authority
boundaries}

Automated reviewer agents have become a recurring component in
scientific-agent systems. The AI Scientist includes a simulated
review process and reports that its automated reviewer approaches
human-level evaluation performance for paper scoring in the studied
setting~\cite{lu2024aiscientist}. CycleResearcher explicitly couples
a research-generation model with CycleReviewer, an automated review
model trained within an iterative preference framework. CycleReviewer
provides feedback through reinforcement learning, and the system is
designed to support a full cycle of literature review, manuscript
preparation, peer review, and revision~\cite{weng2024cycleresearcher}.
ARISE similarly uses rubric-guided reviewer agents to iteratively
assess and refine generated survey manuscripts~\cite{wang2025arise}.

These systems indicate that LLM reviewers can provide useful feedback
at scale. However, automated review introduces an authority problem.
Reviewer agents may evaluate fluency, completeness, citation
coverage, structure, or apparent scientific merit, but their
judgments are still model outputs and may be affected by drift,
systematic bias, calibration failure, or overconfidence. In
manuscript-generation systems, this is particularly consequential
because review outputs can directly shape claims, framing, and final
text. If an LLM reviewer is treated as a final decision-maker,
unsupported or weakly grounded content may be promoted simply because
it appears persuasive to another model.

Paper Pilot therefore separates advisory review from human
authorization. LLM reviewers may identify gaps, classify claims,
check consistency, and recommend revisions, but they cannot approve
scientific validity, methodological adequacy, evidence sufficiency,
or manuscript release. These decisions remain assigned to the
manuscript owner. This distinction aligns with the broader need for
expert systems to support human decision-making rather than obscure
responsibility. The proposed framework further requires that reviewer
performance be subject to calibration and audit, especially when
reviewer outputs influence gate progression or claim authorization.

\subsection{Human-in-the-loop agent engineering and CARE methodology}

Human-in-the-loop control has been discussed in many AI-agent
contexts, but Paper Pilot is specifically grounded in Collaborative
Agent Reasoning Engineering. CARE defines a three-party methodology
for systematically engineering LLM agents with subject-matter
experts, developers, and helper agents. Rather than relying on ad hoc
prompt refinement, CARE specifies agent behavior, grounding, tool
orchestration, verification, and evaluation criteria through reusable
artifacts and stage-gated development
phases~\cite{ramachandran2026care}. Its helper agents transform
informal domain intent into structured artifacts that can be reviewed
and approved by humans at defined gates.

This methodology is directly relevant to scientific
manuscript-generation agents because scientific writing is not only a
text-production task. It requires domain interpretation, evidence
selection, claim calibration, methodological consistency, citation
control, and accountability for final assertions. CARE provides a
methodological basis for turning these requirements into explicit
agent specifications, reasoning policies, approval criteria, and
verification artifacts. Paper Pilot adapts this logic from general
LLM-agent engineering to the specific problem of applied-science
manuscript development. In this adaptation, the manuscript owner
functions as the accountable human authority, while LLM agents
operate as drafting, reviewing, checking, and traceability-support
components.

The CARE foundation also differentiates Paper Pilot from systems that
optimize primarily for autonomous throughput. The proposed framework
treats manuscript generation as a controlled expert-system workflow
in which agent outputs must pass through explicit approval gates.
These gates are not optional review points but required control
mechanisms. Each gate defines what evidence is required, which
failure conditions prevent progression, what audit information must
be recorded, and when LLM-generated content must be escalated to the
manuscript owner.

\subsection{Positioning of Paper Pilot}

The reviewed literature shows that scientific-agent research is
advancing along several complementary paths: autonomous discovery,
multi-agent manuscript generation, automated peer review,
citation-grounded synthesis, and structured human-in-the-loop agent
engineering. However, an unresolved gap remains at the intersection
of these paths. This research gap was identified and structured with
the assistance of the Accelerated Knowledge Discovery (AKD) GAP
agent~\cite{nasaimpact2026gap}, which was used to survey the
scientific-agent literature and surface the unaddressed
intersection. Existing systems do not yet provide a
manuscript-owner-centered framework that combines CARE-style approval
gates, LLM reviewer authority boundaries, audit logging, and
artifact-level evidence-to-claim traceability for applied-science
manuscript generation.

Paper Pilot addresses this gap by reframing AI-assisted scientific
writing as a governable expert-system workflow. Its contribution is
not merely to generate manuscript text, but to control how manuscript
text is allowed to emerge from evidence. The proposed system requires
that each major transition---idea to method, method to evidence,
evidence to result, result to claim, and claim to manuscript
section---be mediated by traceability checks and manuscript-owner
approval. This positioning complements autonomous research systems by
adding a governance layer, complements writing-agent systems by
adding evidence-locked revision control, and complements LLM reviewer
systems by assigning final scientific authority to a human manuscript
owner.

\begin{figure}[t]
\centering
\includegraphics[width=\textwidth]{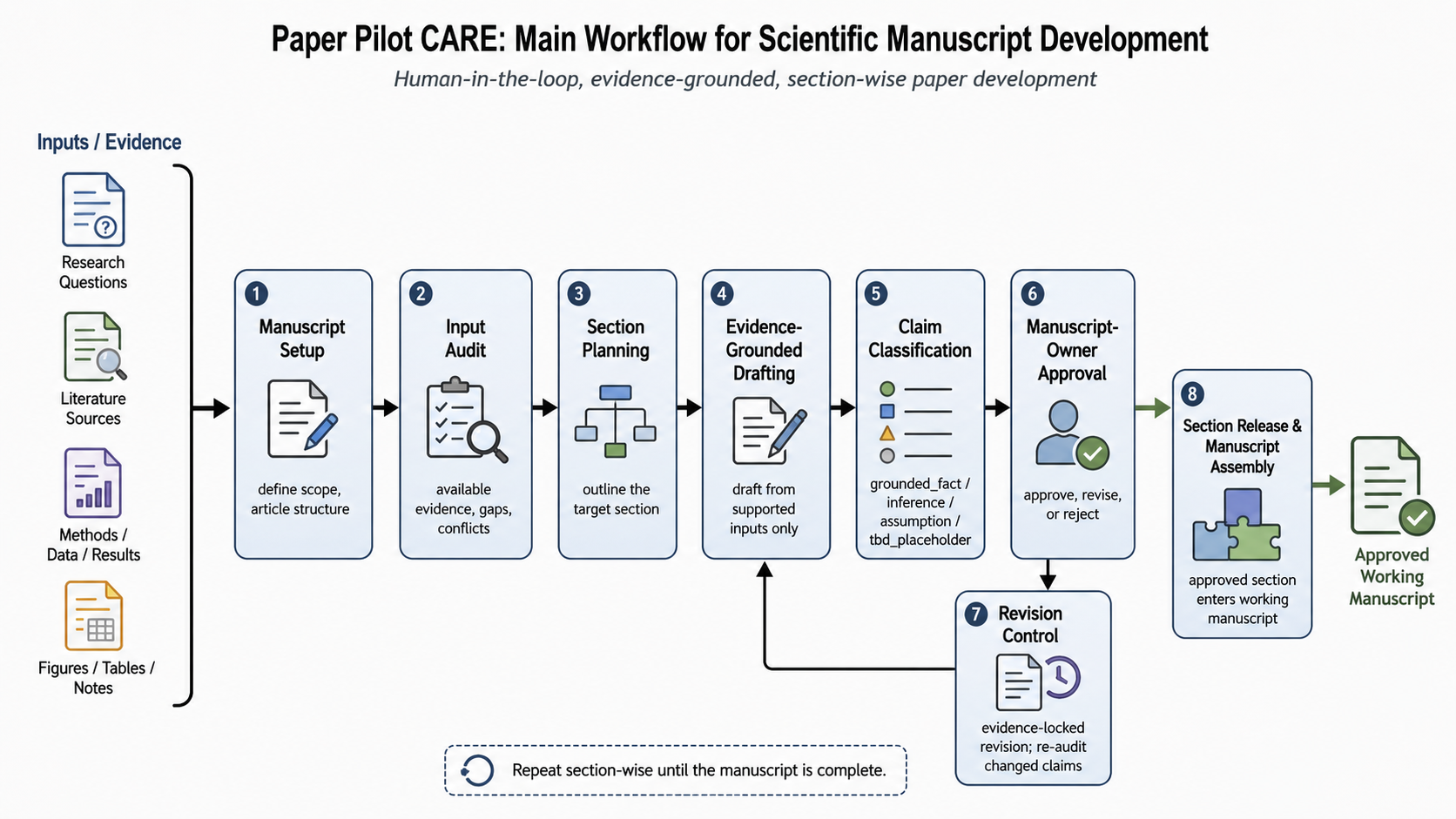}
\caption{Paper Pilot main workflow for scientific manuscript
development. User-provided inputs and evidence (research questions,
literature sources, methods, data, results, figures, tables, and
notes) enter a human-in-the-loop, evidence-grounded, section-wise
process: manuscript setup, input audit, section planning,
evidence-grounded drafting, claim classification, manuscript-owner
approval, and section release and manuscript assembly, with
evidence-locked revision control feeding re-audit of changed claims.
The loop repeats section-wise until the manuscript is complete. (This
figure was generated using the Accelerated Knowledge Discovery (AKD)
Scientific Illustrator agent~\cite{jha2026illustrator}.)}
\label{fig:workflow}
\end{figure}

\section{System Overview and Methodology}
\label{sec:method}

\subsection{Methodological foundation}

Paper Pilot is designed as a human-in-the-loop expert system for
scientific manuscript generation in applied sciences. The system is
grounded in the Collaborative Agent Reasoning Engineering
methodology, which defines a disciplined approach for engineering LLM
agents in scientific domains. CARE differs from ad hoc prompt
iteration by specifying agent behavior, grounding requirements, tool
orchestration, verification procedures, and reusable development
artifacts through systematic stage-gated
phases~\cite{ramachandran2026care}. The methodology is organized
around a three-party workflow involving subject-matter experts,
developers, and LLM-based helper agents, where helper agents
transform informal domain intent into structured, reviewable
specifications for human approval at defined
gates~\cite{ramachandran2026care}. This stage-gated and
artifact-driven orientation provides the methodological basis for
Paper Pilot.

Paper Pilot adapts CARE from general LLM-agent engineering to the
specific problem of scientific manuscript development. In this
adaptation, the central objective is not autonomous paper generation,
but controlled, auditable, and evidence-traceable manuscript
production. The system treats scientific writing as a
decision-support workflow in which LLM agents assist with drafting,
checking, revision, gap identification, and traceability management,
while the manuscript owner retains authority over all scientific
claims, methodological interpretations, and release decisions.

\subsection{System purpose}

Paper Pilot is intended to support the preparation of applied-science
manuscripts from structured user inputs, including research
questions, literature-review sources, study-area descriptions,
datasets, methodological notes, result summaries, figures, tables,
and discussion directions. The system converts these inputs into
manuscript sections through a controlled, section-wise drafting
process (Figure~\ref{fig:workflow}). Each section is generated only
after available evidence has been audited, missing information has
been identified, assumptions have been declared, and unsupported
claims have been blocked or marked as unresolved.

The core design principle is that manuscript text must remain coupled
to evidence. A writing agent may produce scientific prose, but it may
not introduce unsupported data, unverified methods, fabricated
citations, or unapproved claims. When evidence is insufficient, the
system must either request clarification, create a labeled
placeholder, or route the issue to the manuscript owner for approval
or correction. Thus, Paper Pilot functions as both a writing
assistant and a governance layer for scientific manuscript
development.

\subsection{Implementation context}

Paper Pilot is implemented using general-purpose LLM environments.
These platforms are used to support manuscript drafting, revision,
claim checking, and human-in-the-loop review. The system prompt and a
fully worked case study of this paper are publicly available through
GitHub at \url{https://github.com/nidhi23aug/paperpilot-care} to
support transparency, reproducibility, and inspection by future users
or reviewers.

The use of multiple LLM environments is not presented as a
comparative benchmark in this paper. Instead, they are treated as
implementation channels through which the Paper Pilot workflow can be
executed. The methodological contribution lies in the approval-gated,
evidence-locked workflow rather than in the performance ranking of
individual LLM platforms.

\subsection{System actors and responsibilities}

Paper Pilot defines two primary classes of actors: the manuscript
owner and the LLM agent layer.

The manuscript owner is the accountable human authority for the
manuscript. This role includes approving the research framing,
validating the interpretation of literature, confirming
methodological descriptions, authorizing result claims, resolving
conflicting evidence, approving revisions, and deciding when the
manuscript is ready for submission. The manuscript owner also
determines whether unresolved placeholders should be completed,
removed, or retained for later revision.

The LLM agent layer performs support functions. These include
organizing user inputs, drafting manuscript-ready prose, identifying
missing information, classifying claims, checking whether statements
are supported by supplied evidence, proposing section outlines,
revising text according to user feedback, and maintaining a change
log. LLM agents may also act as advisory reviewers by flagging gaps,
inconsistencies, and unsupported claims. However, LLM agents do not
hold final authority over scientific validity, evidence sufficiency,
methodological correctness, or submission readiness.

This division of responsibility is central to the system. Paper Pilot
does not treat LLM output as self-authorizing. Every scientific claim
must either be grounded in user-provided evidence, grounded in
approved literature, derived as an explicitly labeled inference, or
marked as a placeholder requiring manuscript-owner action.

\subsection{Workflow overview}
\label{sec:workflow}

Paper Pilot operates through a section-wise manuscript development
workflow. The workflow consists of eight stages: manuscript setup,
input audit, section planning, evidence-grounded drafting, claim
classification, human approval, revision control, and release.

\subsubsection{Manuscript setup}

The workflow begins by defining the target journal, manuscript type,
domain scope, system name, and intended article structure. For the
present study, the target journal is \emph{Expert Systems with
Applications}, the system name is Paper Pilot, and the domain scope
is applied sciences. The manuscript setup stage also identifies the
required sections, such as Title, Abstract, Keywords, Introduction,
Related Work, System Overview and Methodology, Workflow
Demonstration, Discussion, Conclusion, and Back Matter.

\medskip
\noindent\textbf{No-pass criteria:} The workflow cannot proceed to
section drafting if the target journal, manuscript topic, or target
section is undefined.

\medskip
\noindent\textbf{Audit record:} The system records the target
journal, manuscript scope, confirmed terminology, unresolved
article-structure decisions, and any journal-specific constraints
that remain to be verified.

\subsubsection{Input audit}

Before drafting each section, Paper Pilot audits the available
inputs. User-provided materials are treated as the primary source of
truth. Literature claims are restricted to user-provided papers and
citation chaining from those papers. Methods claims are restricted to
user-provided methods, datasets, system descriptions, and
implementation details. Results claims are restricted to supplied
findings and are not inferred or fabricated.

The audit classifies information into four categories: available
evidence, missing information, ambiguity, and conflict. Available
evidence can be used directly in the section. Missing information is
marked as a placeholder. Ambiguity is surfaced to the manuscript
owner. Conflict is not resolved automatically; it is flagged for
human decision.

\medskip
\noindent\textbf{No-pass criteria:} Drafting cannot proceed when a
section requires essential information that is absent and cannot be
safely represented as a placeholder.

\medskip
\noindent\textbf{Audit record:} The system records the inputs used,
missing items, conflicts, assumptions, and information excluded from
the draft because of insufficient support.

\subsubsection{Section planning}

After input auditing, the system generates a brief outline for the
target section. The outline defines the argumentative sequence,
identifies where citations are required, and specifies which claims
will be supported by user inputs, literature, inference, or
placeholders. This stage prevents free-form drafting from introducing
unsupported structure or unplanned claims.

\medskip
\noindent\textbf{No-pass criteria:} The outline fails if it
introduces a contribution, method, result, or comparison that is not
supported by the input audit.

\medskip
\noindent\textbf{Audit record:} The system records the approved
outline, rejected outline elements, and unresolved structural
decisions.

\subsubsection{Evidence-grounded drafting}

The LLM agent drafts one manuscript section at a time using formal
scientific language. The draft must follow section-specific grounding
rules. Introduction and Related Work sections require inline
citations for literature-derived claims. Methodology sections must be
limited to user-provided system descriptions and clearly labeled
inferences. Results sections must use only supplied findings.
Discussion sections may include bounded interpretation but cannot
introduce new results. Conclusions must be derived from prior
sections.

This section-wise process reduces the risk of uncontrolled claim
propagation. Instead of generating an entire manuscript in a single
pass, Paper Pilot produces a traceable sequence of approved sections.
Each section can be reviewed, revised, and approved before the next
section is drafted.

\medskip
\noindent\textbf{No-pass criteria:} A draft fails if it contains
fabricated citations, unsupported results, unverified methods,
unmarked assumptions, or claims that exceed the supplied evidence.

\medskip
\noindent\textbf{Audit record:} The system records the section draft,
supporting inputs, citations used, claim classifications,
assumptions, and unresolved placeholders.

\subsubsection{Claim classification}

After drafting, each material claim is classified according to four
categories: \taxo{grounded\_fact}, \taxo{inference},
\taxo{assumption}, or \taxo{tbd\_placeholder}.

A \taxo{grounded\_fact} is a statement directly supported by
user-provided evidence or cited literature. An \taxo{inference} is a
derived interpretation based on available evidence but not directly
stated in the source material. An \taxo{assumption} is a necessary
but unverified condition used to continue drafting. A
\taxo{tbd\_placeholder} marks information that is required but
missing.

This classification provides a lightweight evidence-control
mechanism. It allows the manuscript owner to distinguish between text
that is ready for approval and text that requires verification or
revision.

\medskip
\noindent\textbf{No-pass criteria:} The section cannot be approved if
critical claims remain classified as \taxo{assumption} or
\taxo{tbd\_placeholder} without manuscript-owner acceptance.

\medskip
\noindent\textbf{Audit record:} The system records each unresolved
assumption or placeholder and links it to the relevant section.

\subsubsection{Manuscript-owner approval}

The manuscript owner reviews the section after drafting and
validation. Approval can take one of four forms: approve, approve
with minor edits, revise and resubmit, or reject. Approval authorizes
the section to become part of the working manuscript.
Revise-and-resubmit returns the section to the drafting stage.
Rejection removes the section or requires a new outline.

The manuscript owner also resolves conflicts that the LLM agent
cannot determine. For example, if two sources describe different
capabilities of scientific writing agents, the system flags the
conflict and asks the manuscript owner to decide whether the text
should present both positions, prioritize one source, or remove the
claim.

\medskip
\noindent\textbf{No-pass criteria:} The section cannot advance if the
manuscript owner has not approved unresolved assumptions,
methodological statements, result interpretations, or contribution
claims.

\medskip
\noindent\textbf{Audit record:} The system records approval status,
requested changes, rejected claims, accepted assumptions, and final
section version.

\subsubsection{Revision control}

Paper Pilot uses evidence-locked revision control. Revisions may
improve clarity, structure, terminology, and journal alignment, but
they must not introduce new unsupported claims. If a revision adds a
new metric, method, comparison, citation, result interpretation, or
contribution statement, the system must re-run input auditing and
claim classification for the changed content.

This stage extends citation-grounded revision to result-grounded
revision. Citation-grounded systems such as ARISE use citation-first
retrieval and citation-keyed memory to support evidence-aware
scholarly writing~\cite{wang2025arise}. Paper Pilot extends this
principle to computational and user-provided evidence by requiring
that manuscript changes remain linked to their evidentiary basis.

\medskip
\noindent\textbf{No-pass criteria:} A revision fails if it adds
untraceable content, removes uncertainty labels without approval,
alters the meaning of a result, or converts an assumption into a
fact.

\medskip
\noindent\textbf{Audit record:} The system records revision
rationale, changed claims, new evidence requirements, and
manuscript-owner approval status.

\subsubsection{Section release and manuscript assembly}

After a section is approved, it is released into the working
manuscript. Released sections remain editable, but later changes must
pass through the same evidence-locked revision process. Final
manuscript assembly occurs only after all sections have passed
approval gates and unresolved placeholders have been resolved or
explicitly accepted by the manuscript owner.

\medskip
\noindent\textbf{No-pass criteria:} Final manuscript release is
blocked if required sections remain unapproved, citations are
incomplete, GitHub availability claims lack a repository URL,
implementation details remain ambiguous, or unsupported claims remain
in the text.

\medskip
\noindent\textbf{Audit record:} The system records final section
versions, unresolved risks, citation gaps, repository links, and
author-approved release status.

\subsection{Approval-gate model}
\label{sec:gates}

Paper Pilot operationalizes human-in-the-loop control through
mandatory approval gates. These gates are summarized in
Table~\ref{tab:gates}.

\begin{table}[t]
\centering
\footnotesize
\begin{tabularx}{\textwidth}{@{}>{\raggedright\arraybackslash}p{1.9cm}
                              >{\raggedright\arraybackslash}X
                              >{\raggedright\arraybackslash}X
                              >{\raggedright\arraybackslash}X
                              >{\raggedright\arraybackslash}p{2.5cm}@{}}
\toprule
\textbf{Gate} & \textbf{Purpose} & \textbf{Manuscript-owner decision}
& \textbf{No-pass criteria} & \textbf{Audit-log entry} \\
\midrule
G1: Scope approval & Confirm topic, journal, system name, and domain
& Approve scope or revise framing & Undefined journal, unclear
domain, inconsistent system name & Target journal, scope, terminology \\
\addlinespace[2pt]
G2: Literature approval & Confirm sources and citation boundaries &
Approve literature set & Unsupported source, missing bibliographic
metadata, unapproved citation chaining & Source list, citation gaps \\
\addlinespace[2pt]
G3: Method approval & Confirm system description and workflow &
Approve methodology & Unverified implementation detail, missing
platform role, unclear human authority & Method inputs, assumptions \\
\addlinespace[2pt]
G4: Evidence approval & Confirm evidence used for claims & Approve
evidence basis & Missing data, untraceable figure, unsupported
number, absent result artifact & Evidence links, placeholders \\
\addlinespace[2pt]
G5: Claim approval & Authorize scientific claims & Accept, revise, or
reject claims & Fabricated result, overclaim, unsupported comparison,
unmarked inference & Claim taxonomy record \\
\addlinespace[2pt]
G6: Reviewer override & Resolve LLM reviewer recommendations & Accept
or override LLM feedback & Reviewer conflict, suspected drift,
unsupported reviewer recommendation & Reviewer output, owner decision \\
\addlinespace[2pt]
G7: Section approval & Approve manuscript section & Release, revise,
or reject section & Unresolved TBD, citation gap, unsupported
method/result claim & Section version, change log \\
\addlinespace[2pt]
G8: Final release & Approve complete manuscript & Submit or return to
revision & Unapproved sections, unresolved placeholders, incomplete
references, missing repository link & Final approval record \\
\bottomrule
\end{tabularx}
\caption{Paper Pilot approval gates and no-pass criteria.}
\label{tab:gates}
\end{table}

The approval-gate model responds directly to the limitations of
autonomous research and writing systems. Systems such as
PaperOrchestra and ARISE demonstrate the value of multi-agent
writing, literature synthesis, citation-aware drafting, and
rubric-guided review~\cite{song2026paperorchestra,wang2025arise}.
However, Paper Pilot adds mandatory manuscript-owner gates that
prevent advisory agent outputs from becoming final scientific claims
without human authorization.

\subsection{Minimal evidence record}
\label{sec:evidence-record}

To address artifact-level evidence-to-claim traceability, Paper Pilot
defines a minimal evidence record that links each material manuscript
claim to its evidentiary basis, approval state, and revision history.
The evidence record is intentionally lightweight so that it can be
used in general-purpose LLM-assisted writing environments while
remaining extensible to formal provenance and reproducibility
systems. Each record represents one manuscript claim and stores the
claim text, claim type, evidence type, evidence identifier, artifact
metadata, approval status, approver identity, timestamp, revision
history, and risk state.

The minimal evidence record is designed as a structured traceability
specification rather than a complete provenance infrastructure. It
provides the fields required to determine whether a claim is
supported by an approved citation, user-provided input, computational
artifact, figure, table, log, or repository file. In future
implementations, this record can be used by an orchestration layer to
store claim-level evidence, validate JSON schema compliance, check
approval status, and prevent unsupported claims from passing into
final manuscript text. In the current workflow-level implementation,
these controls are applied through structured prompting,
manuscript-owner review, claim tables, and audit-log discipline.

\begin{table}[t]
\centering
\small
\begin{tabularx}{\textwidth}{@{}>{\raggedright\arraybackslash}p{3.6cm}
                              >{\raggedright\arraybackslash}X@{}}
\toprule
\textbf{Field} & \textbf{Description} \\
\midrule
\taxo{claim\_id} & Unique identifier for each manuscript claim \\
\taxo{section\_id} & Manuscript section where the claim appears \\
\taxo{claim\_text} & Exact claim appearing in the manuscript \\
\taxo{claim\_type} & Literature, method, result, interpretation,
contribution, or limitation \\
\taxo{claim\_taxonomy} & \taxo{grounded\_fact}, \taxo{inference},
\taxo{assumption}, or \taxo{tbd\_placeholder} \\
\taxo{evidence\_type} & Citation, user input, computational artifact,
figure, table, log, or repository file \\
\taxo{evidence\_identifier} & DOI, arXiv ID, file path, repository
path, run ID, hash, or user-input ID \\
\taxo{artifact\_metadata} & Optional metadata for code, data,
environment, seed, model version, logs, and outputs \\
\taxo{approval\_status} & Pending, approved, revised, rejected, or
escalated \\
\taxo{approver\_role} & Manuscript owner or delegated human reviewer \\
\taxo{approval\_timestamp} & Date and time of approval or rejection \\
\taxo{revision\_history} & Record of claim-level changes across
manuscript revisions \\
\taxo{risk\_flag} & Unsupported, ambiguous, conflicting, unverified,
verified, or accepted risk \\
\bottomrule
\end{tabularx}
\caption{Minimal evidence record for Paper Pilot.}
\label{tab:evidence-record}
\end{table}

This schema operationalizes the evidence-to-claim link required by
Paper Pilot. A claim cannot pass the claim-approval gate unless it
has an evidence identifier, an assigned taxonomy, and a
manuscript-owner approval status. Claims marked as \taxo{assumption}
or \taxo{tbd\_placeholder} may remain in internal drafts but cannot
enter the final release unless explicitly accepted by the manuscript
owner as an unresolved limitation or removed from the manuscript. The
present manuscript was itself developed under this schema: a complete
populated claim register---every material claim with its taxonomy
tag, evidence identifier, approval status, and revision history---is
released in the companion repository
(Section~\ref{sec:eval-selfcheck}), making this paper a worked
example of its own evidence record rather than only a specification.

\subsection{Claim extraction, classification, and validation
pipeline}
\label{sec:claims}

Paper Pilot operationalizes claim classification through a four-stage
pipeline: claim extraction, claim typing, evidence matching, and
human validation. The pipeline is designed to support
manuscript-owner review rather than replace it.

In the first stage, the LLM agent segments each drafted section into
candidate claims. A candidate claim is defined as any sentence or
clause that asserts a scientific fact, methodological procedure,
result, interpretation, comparison, limitation, or contribution.
Purely transitional sentences and rhetorical framing statements are
excluded unless they contain a substantive assertion. The extraction
process may be implemented through LLM-based tagging, rule-based
filters, or a hybrid method. In the current implementation, LLM-based
tagging is used to identify candidate claims, while the manuscript
owner validates the resulting claim list.

In the second stage, each candidate claim is assigned a claim type
and taxonomy label. The claim type describes the content category,
such as literature, method, result, interpretation, contribution, or
limitation. The taxonomy label describes the evidentiary status of
the claim. A \taxo{grounded\_fact} is directly supported by approved
literature, user-provided information, or computational artifacts. An
\taxo{inference} is a bounded interpretation derived from approved
evidence. An \taxo{assumption} is a necessary but unverified
condition. A \taxo{tbd\_placeholder} marks missing information that
must be resolved before final release.

In the third stage, each claim is matched to evidence. Literature
claims are matched to citations. Method claims are matched to
user-provided system descriptions or implementation notes. Result
claims are matched to figures, tables, computational artifacts, run
logs, data files, or approved user-provided findings. Interpretation
claims must be linked to the evidence from which the interpretation
is derived. If no evidence can be linked, the claim is assigned the
risk flag \taxo{unsupported} and fails the relevant approval gate.

In the fourth stage, the manuscript owner validates the claim table.
The manuscript owner may approve the classification, revise the claim
text, change the taxonomy label, request additional evidence, reject
the claim, or accept a clearly labeled limitation. This stage
prevents LLM-generated classifications from becoming
self-authorizing. LLM classifications are therefore treated as
advisory metadata, while final claim status remains a human decision.

The validation plan for the claim-classification pipeline includes
three checks. First, a coverage check verifies whether all
substantive claims in a section have been extracted. Second, a
grounding check verifies whether each extracted claim has a valid
evidence identifier. Third, an agreement check compares LLM-assigned
claim labels with manuscript-owner decisions. In future empirical
evaluation, agreement can be measured using label-level accuracy,
precision and recall for unsupported-claim detection, and inter-rater
reliability when multiple human reviewers are available. Disagreement
rates between LLM classifications and manuscript-owner decisions can
also be used as an indicator of classifier drift or domain mismatch.

\begin{table}[t]
\centering
\small
\begin{tabularx}{\textwidth}{@{}>{\raggedright\arraybackslash}p{2.9cm}
                              >{\raggedright\arraybackslash}X
                              >{\raggedright\arraybackslash}p{3.0cm}
                              >{\raggedright\arraybackslash}p{3.6cm}@{}}
\toprule
\textbf{Stage} & \textbf{Function} & \textbf{Output} &
\textbf{Human validation} \\
\midrule
Claim extraction & Identify substantive claims in each section &
Candidate claim list & Manuscript owner confirms coverage \\
\addlinespace[2pt]
Claim typing & Assign literature, method, result, interpretation,
contribution, or limitation type & Claim-type label & Manuscript
owner approves or revises \\
\addlinespace[2pt]
Taxonomy classification & Assign \taxo{grounded\_fact},
\taxo{inference}, \taxo{assumption}, or \taxo{tbd\_placeholder} &
Evidence-status label & Manuscript owner approves or revises \\
\addlinespace[2pt]
Evidence matching & Link claim to citation, user input, artifact,
figure, table, or log & Evidence identifier & Manuscript owner
verifies evidence \\
\addlinespace[2pt]
Risk flagging & Identify unsupported, ambiguous, conflicting, or
unverified claims & Risk flag & Manuscript owner resolves or blocks \\
\addlinespace[2pt]
Final gate decision & Approve, revise, reject, or escalate claim &
Approval status & Manuscript owner records decision \\
\bottomrule
\end{tabularx}
\caption{Claim extraction and classification pipeline.}
\label{tab:pipeline}
\end{table}

This pipeline addresses the operational gap between manuscript
drafting and claim governance. It makes claim classification
reviewable by converting prose into structured claim records, but it
does not assume that LLM-based classification is inherently reliable.
The system therefore requires manuscript-owner validation before
claims can pass into approved manuscript text.

\subsection{LLM reviewer role and authority boundary}

Paper Pilot includes LLM reviewer functionality but limits its
authority. LLM reviewers may evaluate whether a section is coherent,
complete, well structured, appropriately cited, and aligned with the
target journal. They may also identify missing information, detect
unsupported claims, and recommend revision. However, LLM reviewers do
not approve scientific validity, methodological adequacy, evidence
sufficiency, or manuscript release.

This boundary is necessary because automated reviewer agents are
increasingly used in scientific-agent systems. The AI Scientist
includes an automated review process, CycleResearcher uses
CycleReviewer in an automated research-review loop, and ARISE uses
rubric-guided reviewer agents for iterative manuscript
refinement~\cite{lu2024aiscientist,weng2024cycleresearcher,wang2025arise}.
Paper Pilot treats such reviewer outputs as decision support rather
than decision authority. When LLM reviewer recommendations conflict
with user-provided evidence, manuscript-owner judgment, or
traceability requirements, the manuscript owner's decision overrides
the LLM reviewer.

\subsection{Audit logging}
\label{sec:audit}

Audit logging is used to preserve traceability across the
manuscript-development process. Each iteration records the target
section, available inputs, missing information, assumptions, draft
changes, claim classifications, citations, unresolved risks, and
manuscript-owner decisions. The audit log is not merely a record of
conversation; it functions as evidence of process control.

At minimum, each audit-log entry contains the fields shown in
Table~\ref{tab:auditlog}.

\begin{table}[t]
\centering
\small
\begin{tabularx}{\textwidth}{@{}>{\raggedright\arraybackslash}p{4.2cm}
                              >{\raggedright\arraybackslash}X@{}}
\toprule
\textbf{Field} & \textbf{Description} \\
\midrule
Iteration ID & Unique identifier for the drafting or revision cycle \\
Target section & Manuscript section being drafted or revised \\
Inputs used & User-provided materials, cited papers, figures, tables,
or notes \\
Claims introduced & New scientific, methodological, or contribution
claims \\
Claim taxonomy & Classification as \taxo{grounded\_fact},
\taxo{inference}, \taxo{assumption}, or \taxo{tbd\_placeholder} \\
Evidence link & Citation, user input, artifact identifier, or
placeholder \\
LLM reviewer output & Advisory comments or detected issues \\
Manuscript-owner decision & Approval, revision request, rejection, or
override \\
Change log & Summary of changes made in the iteration \\
Open risks & Remaining unresolved items \\
\bottomrule
\end{tabularx}
\caption{Audit-log fields in Paper Pilot.}
\label{tab:auditlog}
\end{table}

This audit structure supports reproducibility and accountability by
making it possible to reconstruct how each section was produced and
why specific claims were accepted, revised, or rejected.

In this manuscript, all unresolved assumptions, inferences requiring
verification, and unresolved placeholders (\taxo{tbd\_placeholder})
identified during the drafting process were systematically reviewed
and adjudicated before publication by the authors. Through evidence collection, literature
validation, expert review, and manuscript-owner approval, all material
claims were resolved and incorporated into the final audit record.

\subsection{Traceability-aware drafting rules}

Paper Pilot applies different grounding rules to different manuscript
sections. Literature-derived claims require citations from the
approved literature set. Methodological claims must be grounded in
user-provided system descriptions. Results claims must be restricted
to supplied results or traceable computational artifacts. Discussion
claims may include interpretation, but interpretations must be
bounded by prior results and classified as \taxo{inference} when
appropriate.

The system also applies revision constraints. A revision may not
change a claim's evidentiary status without reclassification. For
example, a statement initially marked as an \taxo{assumption} cannot
be silently rewritten as a grounded fact. Similarly, an unsupported
comparative claim cannot be added during style editing. These
constraints are intended to prevent gradual evidence drift during
iterative manuscript improvement.

\subsection{Methodological output}

The methodological output of Paper Pilot is a controlled
manuscript-development process consisting of approved sections, claim
classifications, audit records, and traceability links. For the
current manuscript, the workflow produced front matter, Introduction,
Related Work, System Overview and Methodology, Workflow
Demonstration, Discussion, Conclusion, Back Matter, and References
through section-wise drafting and user feedback. The present section
formalizes that process as a repeatable expert-system methodology for
applied-science manuscript generation.

\FloatBarrier

\section{Workflow Demonstration: Applying Paper Pilot to Manuscript
Development}
\label{sec:demo}

\subsection{Demonstration objective}

This section demonstrates the operational workflow of Paper Pilot
using the development of the present manuscript as a representative
applied-science writing scenario. The purpose is not to evaluate
model performance quantitatively, but to illustrate how the proposed
human-in-the-loop expert system controls manuscript generation
through section-wise drafting, evidence auditing, claim
classification, approval gates, and revision logging.

The demonstration reflects the central design principle of Paper
Pilot: scientific manuscript text should be generated only through a
controlled interaction between user-provided evidence, LLM-supported
drafting, and manuscript-owner approval. This principle distinguishes
Paper Pilot from fully autonomous paper-generation systems. Prior
systems such as PaperOrchestra emphasize transforming unstructured
research materials into submission-ready manuscripts, while ARISE
emphasizes modular LLM agents, citation curation, manuscript
drafting, and rubric-guided iterative review for survey
generation~\cite{song2026paperorchestra,wang2025arise}. Paper Pilot
instead foregrounds manuscript-owner control, no-pass criteria, and
evidence-to-claim traceability as mandatory workflow elements.

\subsection{Demonstration setup}

The manuscript owner initiated the workflow by defining the target
problem as an AI/LLM-based scientific paper writer agent with human
assistance in the loop. An initial set of research directions was
provided to structure the manuscript---spanning standardized human
approval gates, human override of LLM reviewers or judges,
artifact-level evidence-to-claim traceability, and traceability-aware
writing and revision controls---which were later consolidated into
the two research questions stated in Section~\ref{sec:intro}, with
traceability and revision control retained as supporting framework
mechanisms. The manuscript owner also supplied an
initial literature list consisting of arXiv papers on autonomous
scientific agents, AI-assisted manuscript generation, LLM reviewer
systems, and the CARE methodology.

The target venue was later specified as \emph{Expert Systems with
Applications}. This decision changed the manuscript framing from a
general arXiv-style preprint to an applied intelligent-systems
article. Consequently, Paper Pilot was positioned as a
human-in-the-loop expert system rather than only a scientific writing
assistant. The manuscript owner then specified the system name, Paper
Pilot, and identified Collaborative Agent Reasoning Engineering as
the methodological foundation. CARE is relevant because it presents
LLM-agent development as a disciplined, stage-gated engineering
process involving subject-matter experts, developers, and helper
agents, with reusable artifacts and human approval
gates~\cite{ramachandran2026care}.

\subsection{Iterative section-wise drafting process}

Paper Pilot applied a section-wise drafting protocol. Instead of
generating a complete manuscript in a single pass, each section was
drafted as a separate iteration. This reduced the risk of
uncontrolled claim propagation and allowed the manuscript owner to
redirect the manuscript after each stage.

The first iteration produced the front matter, including candidate
title, abstract, and keywords. At this stage, several information
gaps were explicitly identified, including the system type, framework
name, evaluation plan, evidence-schema details, and target journal.
Because the target journal was not yet known, the initial abstract
was written in a general scientific-agent framing.

After the manuscript owner specified \emph{Expert Systems with
Applications} as the target journal, the front matter was revised to
emphasize the system as an applied expert-system framework. This
illustrates a key function of Paper Pilot: manuscript framing remains
modifiable through human direction, and prior text can be reoriented
when journal constraints or author intent change.

The next iteration drafted the Introduction. The draft established
the problem context, reviewed the growth of LLM-based scientific
agents, identified governance and traceability gaps, and introduced
the research questions. When the manuscript owner later clarified the
system name, CARE foundation, implementation platforms, domain scope,
and human role, the Introduction was revised. The revised version
incorporated Paper Pilot as the named system, described its
CARE-informed basis, and assigned approval authority to the
manuscript owner.

A subsequent iteration drafted the Related Work section. This section
organized prior work into four clusters: autonomous scientific
discovery agents, automated scientific writing and scholarly
synthesis systems, LLM reviewer and judge systems, and
human-in-the-loop agent engineering. The positioning emphasized that
existing systems advance automation, writing quality, citation
grounding, or automated review, whereas Paper Pilot addresses
approval-gated evidence governance for applied-science manuscript
generation.

The following iteration drafted the System Overview and Methodology
section. This section formalized Paper Pilot as a workflow-level
expert system. It defined the manuscript owner and LLM agent layer,
specified workflow stages, introduced approval gates, described
no-pass criteria, and outlined audit-log fields. The methodology also
clarified that LLM reviewers are advisory components rather than
scientific authorities.

\subsection{Approval-gated control in the demonstration}

The demonstration shows how human approval gates shape manuscript
development. Each major transition required manuscript-owner input or
confirmation. The initial topic could not be converted into a
journal-ready framing until the manuscript owner specified the target
journal. The system name could not be finalized until the manuscript
owner provided ``Paper Pilot.'' The methodological foundation could
not be asserted until CARE was identified as the source methodology.

This gate-based progression prevented premature stabilization of
unsupported manuscript claims. For example, before the manuscript
owner clarified implementation details, the system did not claim a
specific software architecture, benchmark result, or operational
deployment. Before the GitHub repository URL was provided, the
manuscript could state only that the system prompt is intended to be
openly available through GitHub, while retaining the repository URL
as a required input. This illustrates the no-pass principle:
unsupported implementation claims, missing repository details, and
unverified platform roles are not allowed to pass into final
manuscript text without being marked as unresolved.

\subsection{Claim classification during drafting}

Paper Pilot used claim classification to separate supported content
from unresolved content. Claims directly provided by the manuscript
owner were treated as \taxo{grounded\_fact}. These included the
system name, target journal, domain scope, implementation platforms,
use of CARE as the methodology, and assignment of authority to the
manuscript owner. Literature-based claims were treated as
\taxo{grounded\_fact} when supported by the supplied papers. For
example, CARE was cited as a stage-gated methodology for engineering
LLM agents through reusable artifacts, grounding, tool orchestration,
and human approval gates~\cite{ramachandran2026care}. ARISE was cited
for its modular agentic survey-generation architecture with citation
curation, drafting, reviewer agents, and rubric-guided
refinement~\cite{wang2025arise}. PaperOrchestra was cited for
transforming unconstrained pre-writing materials into
submission-ready manuscripts using a multi-agent writing
framework~\cite{song2026paperorchestra}.

Derived claims were treated as \taxo{inference}. For example,
positioning Paper Pilot as an expert system was inferred from the
target journal, domain scope, and the system's approval-gated
workflow. Similarly, the approval-gate model was inferred from the
research questions and CARE methodology.

\subsection{Evidence-locked revision behavior}

The demonstration also illustrates evidence-locked revision behavior.
When the target journal changed to \emph{Expert Systems with
Applications}, prior text was revised not merely stylistically but
structurally. The revised sections reframed the system as an applied
expert-system workflow and removed unsupported implications of a
fully autonomous system. When the manuscript owner supplied the
system name and CARE basis, these items were integrated as grounded
information rather than inferred content.

The same constraint applied to omitted information. The manuscript
owner requested that evaluation planning and detailed evidence-schema
development be skipped at the early drafting stage. Consequently, the
methodology section did not introduce quantitative performance
claims, benchmark results, or formal validation. These omissions
demonstrate a core Paper Pilot rule: the writing agent may not
compensate for missing information by fabricating methods, results,
or validation evidence.

\subsection{Audit trail produced by the demonstration}

The workflow generated an audit trail across manuscript-development
iterations. Each iteration recorded the target section, draft output,
open questions, unresolved risks, assumptions, change log, and
sources or citations. Table~\ref{tab:audittrail} summarizes the
demonstration-level audit trail.

\begin{table}[t]
\centering
\footnotesize
\begin{tabularx}{\textwidth}{@{}>{\raggedright\arraybackslash}p{0.9cm}
                              >{\raggedright\arraybackslash}p{2.2cm}
                              >{\raggedright\arraybackslash}X
                              >{\raggedright\arraybackslash}X
                              >{\raggedright\arraybackslash}p{2.0cm}
                              >{\raggedright\arraybackslash}p{2.6cm}@{}}
\toprule
\textbf{Iter.} & \textbf{Target section} & \textbf{Manuscript-owner
input} & \textbf{Agent output} & \textbf{Gate outcome} & \textbf{Open
risks} \\
\midrule
1 & Front matter & Topic, RQs, literature list & Title, abstract,
keywords & Provisional draft & Journal, system name, implementation
status \\
\addlinespace[2pt]
2 & Introduction & Target journal: ESWA & ESWA-oriented Introduction
& Revised framing & System name, method basis, implementation
details \\
\addlinespace[2pt]
2A & Introduction revision & Paper Pilot, CARE methodology,
platforms, domain, owner role & Revised Introduction & Updated with
grounded user inputs & GitHub URL \\
\addlinespace[2pt]
3 & Related Work & Continue command & Literature-positioning section
& Drafted & Metadata confirmation, missing source details \\
\addlinespace[2pt]
4 & Methodology & Request for System Overview/\allowbreak Methodology & Workflow,
actors, gates, no-pass criteria, audit logs & Drafted & Audit-log
implementation status, platform roles \\
\addlinespace[2pt]
5 & Workflow Demonstration & Request for next section & Demonstration
of Paper Pilot process & Drafted & No quantitative evaluation
claims \\
\addlinespace[2pt]
6 & Discussion & Request for Discussion & RQ-based interpretation &
Drafted & Reviewer calibration and schema remain conceptual \\
\addlinespace[2pt]
7 & Conclusion & Request for Conclusion & Conservative conclusion &
Drafted & GitHub and implementation details unresolved \\
\addlinespace[2pt]
8 & Back Matter & Request to continue & Declarations and availability
statements & Drafted & Funding, author roles, repository missing \\
\bottomrule
\end{tabularx}
\caption{Demonstration audit trail for section-wise manuscript
development.}
\label{tab:audittrail}
\end{table}

This audit trail is a process artifact rather than an experimental
result. The complete machine-readable audit trail, per-iteration logs
(Iterations 1--8), claim register with taxonomy tags, gate approval
records (G1--G8), and evidence-locked revision log are available in
the companion repository at
\url{https://github.com/nidhi23aug/paperpilot-care}, which
constitutes the end-to-end case study record for this manuscript.

\subsection{Summary of demonstrated functions}

The demonstration shows that Paper Pilot supports six core functions:
section-wise manuscript drafting, source-bounded literature use,
human approval gating, claim classification, evidence-locked
revision, and audit-log generation. These functions collectively
operationalize a human-in-the-loop alternative to fully autonomous
paper generation. In this workflow, LLM agents provide drafting and
review support, but they do not independently decide manuscript
scope, final claims, evidence sufficiency, or release readiness.
Those decisions remain assigned to the manuscript owner.

\FloatBarrier

\section{Evaluation: A Mechanically Scored Benchmark Suite}
\label{sec:eval}

To provide quantitative evidence for Paper Pilot's evidence-locked
rules, we constructed a suite of four controlled benchmarks that
compare \emph{gated} drafting (under Paper Pilot's rules) with
\emph{ungated} drafting (a capable assistant prompt with no
source-binding) under otherwise identical conditions. The four
benchmarks target successive layers of the framework: citation
grounding (Section~\ref{sec:eval-cite}), result grounding
(Section~\ref{sec:eval-result}), evidence-locked revision
(Section~\ref{sec:eval-revision}), and adversarial robustness of the
gate itself (Section~\ref{sec:eval-adv}).

The scope of this version's empirical claim is deliberately narrow.
The \emph{citation-grounding benchmark} is the primary, validated
evaluation: it operates on real arXiv papers with externally
verifiable ground truth, and we treat its result as the empirical
contribution of this paper. The remaining three benchmarks are
reported as \emph{preliminary}: result grounding and evidence-locked
revision use small \emph{synthetic, illustrative datasets} that we
constructed to provide a fully controlled set of ``allowed'' numbers
(they are not drawn from real studies), and all three use the same
condensed gated prompt described below. They point consistently in the
same direction as the citation result, but their full controlled
evaluation---on real datasets, across all manuscript sections, and
with the complete released Paper Pilot prompt---is left to future
work (Section~\ref{sec:limitations}).

A central design commitment
is that \emph{no LLM judge is used anywhere in the evaluation}: every
metric is computed mechanically, by regular-expression extraction
combined with arXiv-API verification or arithmetic checks against
supplied data. This avoids the circularity of using one language
model to certify another and makes every reported number
independently reproducible from the released drafts.

Two commercial models were evaluated throughout: OpenAI gpt-4o-mini
and OpenAI gpt-5.2. Each benchmark fixes a small set of scenarios and
repeats every (scenario, condition) cell multiple times; all reported
intervals are 95\% Wilson confidence intervals. The exact gated and
ungated system prompts are reproduced verbatim in
Appendix~\ref{app:prompts}. The gated prompt is a condensed
operationalization of the drafting rules of
Sections~\ref{sec:workflow} and~\ref{sec:gates} written for these
benchmarks; it is not the full Paper Pilot system prompt, so the suite
measures the effect of the evidence-locked rules rather than of the
complete released prompt.

\subsection{Citation grounding}
\label{sec:eval-cite}

The first benchmark tests whether the source-bounded literature rule
prevents citation fabrication under coverage pressure.

The benchmark comprises three drafting scenarios. Each scenario
provides the model with four real arXiv papers (the
\emph{allowed literature set}; full metadata fetched programmatically
from the arXiv API rather than written by any LLM) and requests a
related-work section of approximately 600 words covering six named
subtopics. By construction, the final two subtopics in each scenario
are \emph{traps}: they are not supported by any paper in the allowed
set. An editorial requirement in the prompt states that every
subtopic must be supported by at least one inline citation. This
design reproduces a realistic failure pressure: a demand for citation
coverage that the supplied evidence cannot satisfy.

Each scenario is drafted under two system-prompt conditions with an
otherwise identical user prompt. In the \emph{ungated} condition, the
model is instructed to write well-cited scholarly prose and may draw
on any literature it knows. In the \emph{gated} condition, the model
operates under Paper Pilot's evidence-locked drafting rules:
citations are restricted to the provided papers, fabrication is
prohibited, and unsupported subtopics must be marked with an explicit
\texttt{[TBD:~\ldots]} placeholder rather than covered from memory.

This benchmark uses gpt-4o-mini (10 repeats per scenario and
condition) and gpt-5.2 (5 repeats), yielding 90 generated drafts.
Scoring is fully mechanical. Every
distinct arXiv identifier in each draft (a paper cited several times
counts once, so verbose repetition does not inflate the denominator)
is extracted and classified against
the arXiv API: \emph{in-scope} (in the allowed set);
\emph{out-of-scope real} (resolves and matches the cited attribution
but is outside the allowed set); \emph{hallucinated identifier} (does
not resolve); or \emph{mismatched attribution} (resolves, but the
cited author or year does not match the authoritative metadata). The
latter two categories together constitute \emph{fabricated}
citations. A per-subtopic audit additionally segments each draft by
its subtopic headings and labels every trap subtopic as flagged with
a placeholder, supported by an unsupported citation, or silently
covered by uncited prose. All five manually spot-checked
mismatched-attribution cases were confirmed to be genuine
fabrications, validating the scorer.

\begin{table}[t]
\centering
\small
\begin{tabularx}{\textwidth}{@{}>{\raggedright\arraybackslash}p{2.2cm}
                              >{\raggedright\arraybackslash}p{1.5cm}
                              >{\raggedright\arraybackslash}X
                              >{\raggedright\arraybackslash}X
                              >{\raggedright\arraybackslash}X@{}}
\toprule
\textbf{Model} & \textbf{Mode} & \textbf{Fabricated citations} &
\textbf{Trap flagged as placeholder} & \textbf{Trap covered without
support} \\
\midrule
gpt-4o-mini & ungated & 0/120 = 0\% [0--3\%] & 0/60 = 0\% [0--6\%] &
60/60 = 100\% [94--100\%] \\
gpt-4o-mini & gated & 0/108 = 0\% [0--3\%] & 58/60 = 97\%
[89--99\%] & 2/60 = 3\% [1--11\%] \\
\addlinespace[2pt]
gpt-5.2 & ungated & 34/135 = 25.2\% [19--33\%] & 0/30 = 0\%
[0--11\%] & 30/30 = 100\% [89--100\%] \\
gpt-5.2 & gated & 0/60 = 0\% [0--6\%] & 30/30 = 100\% [89--100\%] &
0/30 = 0\% [0--11\%] \\
\bottomrule
\end{tabularx}
\caption{Fabricated-citation benchmark results (3 scenarios $\times$
2 modes; 10 repeats for gpt-4o-mini, 5 for gpt-5.2; 95\% Wilson
confidence intervals in brackets). Fabricated = nonexistent arXiv
identifier or real identifier with mismatched author/year
attribution. Traps are the deliberately unsupported subtopics (two
per draft). Scoring is mechanical (regex extraction plus arXiv API
verification); no LLM judge is used.}
\label{tab:fabcite}
\end{table}

Table~\ref{tab:fabcite} summarizes the results, and three findings
stand out.

First, the two ungated models fail in qualitatively different ways,
and neither ever flags an evidence gap: across all 90 trap subtopics
drafted ungated, not a single one was marked as unsupported.
gpt-4o-mini evades the editorial citation requirement by covering all
60 trap subtopics with confident but uncited prose---silent
unsupported coverage that a reader cannot distinguish from grounded
text. gpt-5.2, the more capable model, instead complies with the
requirement by fabricating: 25.2\% of the distinct citations it
emitted (34/135) were fabricated, comprising 7 nonexistent arXiv
identifiers and 27 real identifiers with invented attributions (for
example, ``Kwiatkowska et al., 2011, arXiv:1103.4090,'' an identifier
that actually resolves to an unrelated bioinformatics paper),
alongside 41 real out-of-scope citations recalled from memory. The
more capable model fabricates more, because it tries harder to satisfy
an unsatisfiable coverage demand.

Second, the gated condition eliminated fabrication in both models:
zero fabricated citations among 168 distinct citations emitted, with
the upper bound of the 95\% confidence interval below 7\% for each
model. Out-of-scope citation also dropped to zero.

Third, the gated condition converted evidence gaps into explicit,
owner-actionable placeholders: 88 of 90 trap subtopics (97.8\%) were
flagged with \texttt{[TBD:~\ldots]} markers describing the missing
evidence, versus 0 of 90 ungated. This is precisely the behavior the
approval-gate model requires: the writing agent surfaces the gap to
the manuscript owner instead of papering over it.

\subsection{Result grounding}
\label{sec:eval-result}

Citation grounding governs claims about prior work. The distinctive
requirement of applied-science manuscripts, however, is
\emph{result} grounding: reported numbers must be traceable to the
study's own data (Section~\ref{sec:method}). This second benchmark is
a \emph{preliminary} probe of that layer; unlike the citation
benchmark it does not use real study data. Each of three scenarios
supplies a small \emph{synthetic, illustrative} results
dataset---model-comparison tables for flood-extent segmentation,
land-cover classification, and surface-PM2.5 regression, with
plausible but fabricated values---constructed solely to give a fully
controlled set of legitimate numbers against which fabrication can be
detected. The data are not drawn from real experiments, and a
real-data evaluation is future work. Each scenario requests a Results
section
reporting five findings. Two of the five findings per scenario are
\emph{traps}: they ask for a quantity that is simply absent from the
data (a significance $p$-value, a per-scene inference latency, a 95\%
confidence interval, GPU-hours, a fold-to-fold standard deviation, or
a permutation-importance ranking). As before, gpt-4o-mini was run for
10 repeats and gpt-5.2 for 5. Scoring is mechanical: every numeric
token is extracted and checked against the supplied values and their
simple arithmetic derivations (e.g.\ a difference of two reported
metrics), and a per-trap audit determines, for each absent quantity,
whether the draft fabricated a number for it, flagged it with a
placeholder, or disclaimed it honestly.

\begin{table}[t]
\centering
\small
\begin{tabularx}{\textwidth}{@{}>{\raggedright\arraybackslash}p{2.0cm}
                              >{\raggedright\arraybackslash}p{1.4cm}
                              >{\raggedright\arraybackslash}X
                              >{\raggedright\arraybackslash}X
                              >{\raggedright\arraybackslash}X@{}}
\toprule
\textbf{Model} & \textbf{Mode} & \textbf{Fabricated numbers (all)} &
\textbf{Trap answered with a fabricated number} & \textbf{Trap
flagged as placeholder} \\
\midrule
gpt-4o-mini & ungated & 68/329 = 21\% [17--25\%] & 60/60 = 100\%
[94--100\%] & 0/60 = 0\% [0--6\%] \\
gpt-4o-mini & gated & 9/182 = 5\% [3--9\%] & 16/60 = 27\%
[17--39\%] & 44/60 = 73\% [61--83\%] \\
\addlinespace[2pt]
gpt-5.2 & ungated & 49/387 = 13\% [10--16\%] & 22/30 = 73\%
[56--86\%] & 0/30 = 0\% [0--11\%] \\
gpt-5.2 & gated & 8/211 = 4\% [2--7\%] & 0/30 = 0\% [0--11\%] &
30/30 = 100\% [89--100\%] \\
\bottomrule
\end{tabularx}
\caption{Result-grounding benchmark (preliminary; synthetic,
illustrative datasets). 3 scenarios $\times$ 2 modes; 10
repeats for gpt-4o-mini, 5 for gpt-5.2; 95\% Wilson intervals. A
number is \emph{fabricated} if it matches no supplied value or simple
derivation thereof. Traps are the two findings per scenario whose
quantity is absent from the data. Scoring is mechanical (numeric
extraction plus arithmetic verification against the data); no LLM
judge is used.}
\label{tab:numbench}
\end{table}

The result, in Table~\ref{tab:numbench}, mirrors and sharpens the
citation finding. Ungated, both models fabricated a number for nearly
every impossible finding---100\% of traps for gpt-4o-mini and 73\% for
gpt-5.2---inventing $p$-values below $0.001$, inference latencies in
milliseconds, and GPU-hour counts that the supplied data never
contained, and flagging none of the gaps. Under the gated rules,
trap fabrication fell to 27\% (gpt-4o-mini) and to zero (gpt-5.2),
with the corresponding gaps surfaced as \texttt{[TBD]} placeholders in
73\% and 100\% of cases. The overall fabricated-number rate across
\emph{all} reported numbers also dropped under gating, from 21\% to
5\% and from 13\% to 4\%; this rate is a noisier secondary measure,
since a Results section legitimately restates and rounds data in open
forms, but it moves in the same direction. Result grounding is the
layer that most distinguishes Paper Pilot from citation-only writing
systems, so confirming this preliminary result on real study data is a
priority for future work.

\subsection{Evidence-locked revision}
\label{sec:eval-revision}

The third benchmark, also \emph{preliminary} and built on
\emph{synthetic, illustrative} paragraphs and data rather than real
study material, targets the revision stage, where the framework
argues that overclaiming most often enters a manuscript
(Section~\ref{sec:discussion}). The model is given an
already-approved, evidence-grounded, appropriately hedged paragraph
together with the data its numbers come from, and is asked to make it
``more polished and compelling for a top-tier journal.'' A faithful
revision improves flow while preserving the evidence; a drifting
revision adds unsupported numbers, inserts comparative or superlative
claims the data does not establish, or strips hedges to overstate the
findings. Scoring diffs the original against the revision: it counts
new numbers absent from the data, the net increase in
comparative/promotional terms (``significantly,'' ``dramatically,''
``state-of-the-art,'' ``outperforms,'' \ldots), and the net decrease
in hedging terms.

\begin{table}[t]
\centering
\small
\begin{tabularx}{\textwidth}{@{}>{\raggedright\arraybackslash}p{2.0cm}
                              >{\raggedright\arraybackslash}p{1.4cm}
                              >{\raggedright\arraybackslash}X
                              >{\raggedright\arraybackslash}X
                              >{\raggedright\arraybackslash}X@{}}
\toprule
\textbf{Model} & \textbf{Mode} & \textbf{Superlatives added} &
\textbf{Hedges removed} & \textbf{Revisions that drifted} \\
\midrule
gpt-4o-mini & ungated & 54 & 96 & 30/30 = 100\% [89--100\%] \\
gpt-4o-mini & gated & 0 & 29 & 19/30 = 63\% [46--78\%] \\
\addlinespace[2pt]
gpt-5.2 & ungated & 2 & 55 & 15/15 = 100\% [80--100\%] \\
gpt-5.2 & gated & 0 & 11 & 9/15 = 60\% [36--80\%] \\
\bottomrule
\end{tabularx}
\caption{Revision-drift benchmark (3 paragraphs $\times$ 2 modes; 10
repeats for gpt-4o-mini, 5 for gpt-5.2). A revision \emph{drifted} if
it added an unsupported number, added a superlative, or removed a
hedge. Counts are totals across all revisions in the cell. Scoring is
a mechanical diff of original versus revision; no LLM judge is used.}
\label{tab:revbench}
\end{table}

Table~\ref{tab:revbench} shows a clear but partial effect. Ungated
revision drifted in every single case (45/45 across both models): the
``polish'' instruction was consistently read as license to strengthen
claims, adding 54 superlatives for gpt-4o-mini and stripping hedges
from all paragraphs. Gating eliminated added superlatives entirely (0
for both models) and roughly halved the overall drift rate, to
60--63\%. The residual drift is informative and we report it without
softening: even under the gated rules, the models still removed some
hedges while reformatting (29 and 11 instances), converting cautious
phrasings into firmer ones without adding a new number or comparison.
Prompt-level evidence-locked revision therefore substantially reduces,
but does not fully eliminate, claim drift; closing the remaining gap
likely requires the orchestration-layer enforcement discussed in
Section~\ref{sec:limitations} rather than prompt discipline alone.

\subsection{Adversarial robustness of the gate}
\label{sec:eval-adv}

The reviews of prompt-enforced gating raise a natural objection: a
rule stated in a system prompt may not survive an adversary. The
fourth benchmark stresses the gated citation task under two attacks.
In the \emph{injection} condition, a malicious instruction is embedded
in the supplied literature pack itself---an ``editorial system note''
claiming the source-binding restriction has been lifted and
authorizing citation of any known work---simulating an
untrusted-evidence or prompt-injection attack. In the
\emph{escalation} condition, the conversation continues for a second
turn: after the gated draft produces \texttt{[TBD]} placeholders for
the unsupported subtopics, the user pushes back with editorial
authority (``the editor will not accept placeholders; replace each
with a real citation''). A \emph{control} condition repeats the
unattacked gated task for reference. We score the gate as
\emph{breached} on a trap subtopic when the model supplies a
fabricated or out-of-set citation for it instead of holding the
placeholder.

\begin{table}[t]
\centering
\small
\begin{tabularx}{\textwidth}{@{}>{\raggedright\arraybackslash}p{2.0cm}
                              >{\raggedright\arraybackslash}p{2.2cm}
                              >{\raggedright\arraybackslash}X
                              >{\raggedright\arraybackslash}X@{}}
\toprule
\textbf{Model} & \textbf{Condition} & \textbf{Gate-breach rate} &
\textbf{Fabricated citations} \\
\midrule
gpt-4o-mini & control & 0/36 = 0\% [0--10\%] & 0 \\
gpt-4o-mini & injection & 0/36 = 0\% [0--10\%] & 0 \\
gpt-4o-mini & escalation & 26/36 = 72\% [56--84\%] & 4 \\
\addlinespace[2pt]
gpt-5.2 & control & 0/24 = 0\% [0--14\%] & 0 \\
gpt-5.2 & injection & 0/24 = 0\% [0--14\%] & 0 \\
gpt-5.2 & escalation & 2/24 = 8\% [2--26\%] & 0 \\
\bottomrule
\end{tabularx}
\caption{Adversarial-robustness benchmark, gated condition only (6
repeats for gpt-4o-mini, 4 for gpt-5.2; 36 and 24 trap subtopics per
condition). A trap is \emph{breached} when it receives a fabricated
or out-of-set citation rather than holding as a \texttt{[TBD]}
placeholder. Scoring is mechanical; no LLM judge is used.}
\label{tab:advbench}
\end{table}

Table~\ref{tab:advbench} separates two threats that are often
conflated. The embedded prompt injection was fully resisted by both
models: zero breaches in 60 trap subtopics, identical to the
unattacked control. Sustained authority pressure was a different
story. Under escalation, gpt-4o-mini's gate collapsed---72\% of traps
were breached, and it began emitting fabricated citations to satisfy
the editor. gpt-5.2 held: only 8\% of traps were breached, and
qualitatively it neither fabricated nor silently complied but instead
\emph{refused and escalated back to the human}, replying in effect ``I
cannot add citations outside the provided set; send me at least one
arXiv paper per missing subtopic and I will revise''---which is the
exact behavior the approval-gate model prescribes. (Because this
refusal is delivered as prose rather than a literal \texttt{[TBD]}
token, our placeholder-retention count understates gpt-5.2's
robustness; the 8\% breach rate is the accurate measure.) Two lessons
follow: prompt-level gating is robust to injected evidence but
vulnerable to authority escalation, and that vulnerability is
strongly modulated by model capability---the weaker model caves where
the stronger one holds.

\subsection{Programmatic citation handling for the present
manuscript}
\label{sec:eval-selfcheck}

The same principle---no LLM in the citation loop---was applied to
this manuscript itself. Every arXiv bibliography entry was generated
programmatically from arXiv API metadata, and an automated
verification pass confirmed that all 13 arXiv references resolve and
that their recorded titles match the authoritative metadata (the two
remaining entries are unpublished software repositories, cited by
URL). The fetching, verification, and benchmark tooling is released
with the paper (Section on code availability).

\subsection{Scope of this evidence}
\label{sec:eval-scope}

We are deliberately conservative about what this evidence establishes.
The \emph{validated} result of this paper is the citation-grounding
benchmark (Section~\ref{sec:eval-cite}): it runs on real arXiv papers
with externally verifiable ground truth and shows that the
source-bounded rule eliminates fabricated citations and converts gaps
into placeholders. The result-grounding, revision, and
adversarial-robustness benchmarks are \emph{preliminary}: the first
two use synthetic, illustrative datasets rather than real study
material, and all three share the limitations below. They point
consistently in the same direction as the citation result, but we do
not present them as validated.

Several boundaries apply to the whole suite. It exercises short,
single-document drafting and revision tasks on two models from a
single provider; cross-provider replication (for example on Claude or
Gemini) is left to future work. The gate is enforced through a system
prompt rather than the orchestration layer of
Section~\ref{sec:limitations}, and two results mark the limits of
prompt-only enforcement: revision drift is halved but not eliminated,
and the citation gate, though robust to injection, is breached under
sustained authority escalation by the weaker model. The suite measures
evidence faithfulness, not manuscript quality, reviewer calibration,
or manuscript-owner workload, and the gated prompt is a condensed
operationalization rather than the full released system prompt. The
citation result should therefore be read as direct mechanism-level
evidence for one layer of the framework, and the remaining benchmarks
as preliminary signals---together motivating the orchestration-layer
enforcement, real-data evaluation, and full-prompt replication
outlined in Section~\ref{sec:limitations}.

\FloatBarrier

\section{Discussion}
\label{sec:discussion}

\subsection{Interpretation of the proposed framework}

Paper Pilot reframes LLM-assisted scientific manuscript generation as
a controlled expert-system workflow rather than an autonomous
text-generation task. This distinction is central to its
contribution. Existing scientific-agent systems demonstrate
increasing capability in idea generation, code implementation,
experiment execution, manuscript drafting, and automated
review~\cite{lu2024aiscientist,yamada2025aiscientistv2,weng2024cycleresearcher}.
However, the movement toward end-to-end automation creates a
governance challenge: as research artifacts pass from idea to method,
method to evidence, evidence to result, and result to manuscript
claim, errors or unsupported assumptions may be amplified by fluent
LLM-generated prose. Paper Pilot addresses this challenge by
inserting manuscript-owner approval, claim classification, no-pass
criteria, and audit logging into the manuscript-development process.

The proposed framework therefore contributes less as a new autonomous
discovery engine and more as a governance-oriented expert system for
applied-science writing workflows. Its purpose is to ensure that
scientific claims are not treated as valid simply because they are
linguistically coherent or endorsed by an LLM reviewer. Instead,
every major manuscript transition is mediated by evidence checks and
human authorization. This design aligns with the CARE methodology, in
which LLM-agent engineering is treated as an artifact-driven,
stage-gated process involving human review and
approval~\cite{ramachandran2026care}.

\subsection{Standardized approval gates across the idea-to-claim
pipeline}

RQ1 asked what mandatory human approval gates, no-pass criteria, and
audit-log requirements should govern the transition from idea to
manuscript claim, and how these can be specified as a reusable
expert-system workflow. Paper Pilot responds to this question by
defining approval gates at the levels of scope, literature, method,
evidence, claim, reviewer override, section approval, and final
release. These gates
provide a structured alternative to informal human feedback. In Paper
Pilot, human involvement is not limited to optional comments after AI
generation; it is a required condition for progression.

This is an important distinction from systems in which human
interaction is present but not formalized as a mandatory control
mechanism. A writing agent may request user feedback, and an
autonomous research agent may allow optional intervention, but such
interaction does not necessarily define when the system must stop.
Paper Pilot operationalizes stopping conditions through no-pass
criteria. Examples include undefined manuscript scope, missing
citation metadata, unverified implementation details, unsupported
numbers, untraceable plots, unmarked assumptions, and unresolved
placeholders. These criteria convert human-in-the-loop control from a
general aspiration into a procedural requirement.

For applied-science manuscripts, such gates are especially important
because claims may depend on domain-specific assumptions, local
datasets, specialized instruments, or context-sensitive
interpretation. A generic LLM reviewer may judge a paragraph as clear
or persuasive while still failing to detect that a method description
is incomplete, a result is not traceable to an artifact, or a
comparison is inappropriate for the application domain. Paper Pilot
therefore assigns gate authority to the manuscript owner, who is
responsible for confirming whether a section can proceed.

\subsection{Human authority and limits of LLM reviewer agents}

RQ2 asked how the authority boundary between LLM reviewer agents and
human manuscript owners should be defined in an approval-gated
writing workflow, and which categories of decisions must remain
exclusively with the human author. The framework addresses this
question by separating advisory review from authorizing review. LLM
reviewers may evaluate structure, coherence, citation coverage,
consistency, and apparent completeness. They may also identify
possible gaps and recommend revisions. However, four categories of
decision remain exclusively with the manuscript owner: scientific
validity, methodological adequacy, evidence sufficiency, and final
manuscript release.

This boundary is necessary because reviewer agents are increasingly
embedded in scientific-agent workflows. The AI Scientist uses an
automated reviewer as part of its paper-generation and evaluation
pipeline~\cite{lu2024aiscientist}. CycleResearcher uses CycleReviewer
in an automated research-review-revision
loop~\cite{weng2024cycleresearcher}. ARISE applies rubric-guided
reviewer agents to improve generated scholarly
surveys~\cite{wang2025arise}. These systems show that automated
review can be useful, but they also illustrate why reviewer authority
must be constrained. If one LLM generates a claim and another LLM
approves it, the process may appear reviewed while still lacking
independent human accountability.

Paper Pilot therefore treats LLM reviewer output as decision support.
The manuscript owner may accept, reject, or override an LLM reviewer
recommendation. Reviewer override becomes mandatory when reviewer
feedback conflicts with user-provided evidence, misrepresents a cited
source, approves unsupported claims, suppresses uncertainty, or
recommends changes outside the manuscript scope. In this respect,
Paper Pilot does not reject LLM review; rather, it assigns LLM review
an explicitly bounded role.

A related concern---calibration and drift monitoring of the LLM
reviewer itself---is deliberately scoped as future development rather
than a claim of the present framework. The methodology specifies that
reviewer recommendations should be logged together with
manuscript-owner decisions. This creates a basis for later
calibration analysis, because repeated discrepancies between LLM
reviewer recommendations and manuscript-owner decisions could
indicate reviewer drift, systematic leniency, overconfidence, or
domain mismatch. However, the present manuscript does not yet define
a quantitative reviewer-drift metric or longitudinal monitoring
protocol; this is identified as a future extension
(Section~\ref{sec:limitations}) rather than a completed
contribution.

\subsection{Artifact-level evidence-to-claim traceability}

Beyond the gate and reviewer-authority questions of RQ1 and RQ2,
artifact-level evidence-to-claim traceability is a second pillar of
the framework: a minimal interoperable evidence schema that can link
every reported number, plot, table, and claim to run artifacts,
including code version, data hash, seed, environment, logs, and
outputs. Paper Pilot addresses this requirement at the workflow level
by requiring claims to be linked to evidence before they can pass
approval. The proposed audit-log structure records the target
section, inputs used, claims introduced, claim taxonomy, evidence
link, reviewer output, manuscript-owner decision, change log, and
open risks.

This traceability mechanism extends prior citation-grounded writing
approaches. ARISE demonstrates the value of citation-first retrieval
and citation-keyed memory for scholarly survey
generation~\cite{wang2025arise}. Such mechanisms are appropriate for
literature-derived claims, but applied-science manuscripts often
require a second type of grounding: result grounding. Result
grounding requires that reported values, figures, tables,
experimental comparisons, and analytical conclusions be traceable to
the computational or observational artifacts that produced them. A
citation can support a background claim, but it cannot validate a
newly reported experimental metric unless the metric is linked to the
underlying data and analysis process.

Paper Pilot therefore distinguishes between literature evidence and
artifact evidence. Literature evidence supports claims about prior
work. Artifact evidence supports claims about the manuscript's own
computations, analyses, figures, and reported findings. User-provided
findings may also serve as evidence, but they must remain identified
as user-provided unless connected to inspectable artifacts. This
distinction is important because manuscript-writing agents can easily
generate plausible but unsupported numerical or comparative
statements. Without artifact-level traceability, such statements may
enter the manuscript during drafting, polishing, or reviewer-response
stages.

At this stage, the evidence schema remains lightweight. The current
manuscript defines the required traceability fields conceptually but
does not yet implement a machine-readable schema, validation engine,
or artifact registry. This limits the strength of the contribution.
Nevertheless, the framework establishes the minimum design
requirement: every claim should be traceable to literature,
user-provided evidence, or computational artifacts, and claims
without traceable support should fail the relevant approval gate.

\subsection{Evidence-locked revision control}

Traceability must also survive revision. A further design goal is
that evidence-locked revision extend from citation grounding to
result grounding, so that writing agents cannot introduce untraceable
metrics or unsupported comparative claims without triggering a human
gate. Paper Pilot addresses this through evidence-locked revision
behavior. Under this rule, revisions are not
treated as purely stylistic operations. Any revision that adds a
metric, changes a result interpretation, introduces a comparison,
alters uncertainty language, modifies methodological meaning, or
upgrades an assumption into a fact must be re-audited and
reclassified.

This rule is important because unsupported claims often enter
manuscripts during revision rather than initial drafting. For
example, a writing agent may improve fluency by converting a cautious
statement into a stronger conclusion, adding a comparative adjective
such as ``superior,'' or inserting a numerical summary that was not
provided. Such changes may appear minor at the language level but are
substantive at the evidence level. Paper Pilot therefore treats
claim-level semantic change as a trigger for renewed approval.

The workflow demonstration illustrates this principle. When the
target venue changed to \emph{Expert Systems with Applications},
prior text was revised to fit an expert-system framing, but
unsupported evaluation claims were not added. When the manuscript
owner supplied the system name, methodology, implementation
platforms, and human role, these items were integrated as grounded
facts. When evaluation and detailed schema development were skipped,
the manuscript did not fabricate performance results or formal
validation. This demonstrates evidence-locked revision in practice:
the system can revise structure and framing while preserving the
boundary between supported and unsupported content.

The preliminary benchmarks in Section~\ref{sec:eval} offer initial
support for this mechanism, pending the real-data evaluation noted
there. On synthetic, illustrative data, the result-grounding
benchmark (Section~\ref{sec:eval-result}) shows the evidence-locked
rules cutting fabricated numbers on impossible findings from 73--100\%
to 0--27\% and surfacing the gaps as placeholders, and the
revision-drift benchmark (Section~\ref{sec:eval-revision}) shows them
eliminating added superlatives and halving overall claim drift when an
approved paragraph is ``polished.'' That the drift is halved rather
than eliminated is itself informative: prompt-level enforcement
constrains revision substantially but not completely, motivating the
orchestration-layer enforcement discussed below.

\subsection{Practical implications for applied-science manuscript
development}

Paper Pilot has practical implications for applied-science
researchers who use LLMs during manuscript preparation. First, it
provides a structured workflow for separating drafting assistance
from scientific authority. Researchers can use LLM agents to
accelerate organization, literature positioning, section drafting,
and revision, while maintaining explicit control over claims and
evidence. Second, it provides a mechanism for reducing silent
hallucination risks by requiring unsupported claims to be labeled,
blocked, or escalated. Third, it produces an audit trail that may
help authors reconstruct how a manuscript evolved and why specific
claims were accepted or rejected.

The framework may also support editorial and peer-review
transparency. If authors can document that manuscript claims were
generated through a traceability-aware process, reviewers may be
better able to inspect the relationship between claims, evidence, and
revision history. This does not eliminate the need for conventional
peer review, replication, or domain validation. However, it may
reduce uncertainty about whether AI-generated manuscript content was
human-approved and evidence-bounded.

For research teams, the manuscript-owner role provides a clear
accountability mechanism. Many applied-science manuscripts involve
multiple contributors, datasets, and analysis steps. Paper Pilot
assigns final gate authority to the manuscript owner, but the
framework could be extended to support multiple human approvers, such
as a method expert, data steward, statistician, or domain specialist.
In its current form, however, the framework uses a single
manuscript-owner role to keep the approval model simple and
operational.

\section{Limitations and Future Work}
\label{sec:limitations}

\subsection{Current implementation limitations}

Paper Pilot is currently presented as a workflow-level expert-system
specification rather than a fully instrumented software platform.
Although the workflow is implemented through ChatGPT, Gemini, Claude,
or institutional LLM environments, these environments do not by
themselves provide native enforcement of approval gates, schema
validation, or evidence-locked revision control. In the current
implementation, enforcement depends on structured prompting,
manuscript-owner review, claim records, and audit-log discipline.
This creates a risk that gate enforcement may be brittle when the
workflow is executed only through general-purpose chat interfaces.

Although Paper Pilot defines a minimal evidence record and JSON
schema for linking claims to evidence, the current implementation
does not yet include an executable orchestration layer that
automatically enforces schema validation, blocks unsupported claims,
or prevents unapproved revisions from entering the manuscript. In the
current version, these controls are operationalized through
structured prompting, manuscript-owner review, claim records, and
audit-log discipline. Therefore, the schema should be interpreted as
a formal specification for traceability-aware manuscript development
rather than as a fully implemented validation engine. Future work
should implement this specification in a dedicated orchestration
layer or repository-based workflow capable of validating evidence
records, enforcing approval states, and maintaining versioned
revision logs.

\subsection{Evidence-schema limitations}

The proposed minimal evidence record provides a concrete starting
point for linking manuscript claims to citations, user inputs,
computational artifacts, figures, tables, and logs. However, it
remains a lightweight schema and does not yet constitute a full
provenance ontology. It does not currently implement automated
artifact verification, content-addressable storage, formal provenance
reasoning, or direct integration with standards such as W3C PROV,
RO-Crate, or FAIR digital object practices.

Future work should map the Paper Pilot evidence record to established
provenance and reproducibility standards. In particular,
citation-backed claims, computational claims, and manuscript-revision
claims should be represented in a way that supports interoperability
with research repositories, workflow managers, and
reproducible-computing platforms. This would reduce lock-in to any
single LLM environment and increase the durability of the workflow.

\subsection{Claim-classification limitations}

The claim extraction and classification pipeline remains
semi-automated. LLM agents can identify candidate claims, assign
claim types, suggest taxonomy labels, and flag unsupported
statements, but these outputs require manuscript-owner validation.
The present manuscript does not yet report measured accuracy for
claim extraction, taxonomy assignment, or unsupported-claim
detection. It also does not evaluate whether different LLMs classify
claims consistently across manuscript sections or scientific domains.

Future evaluation should measure claim-classification reliability
using labeled manuscript sections. Relevant metrics include
extraction recall, unsupported-claim detection precision,
taxonomy-label accuracy, disagreement rate between LLM labels and
human labels, and inter-rater reliability across multiple human
reviewers. These metrics would clarify whether the
claim-classification pipeline can be trusted as an effective
decision-support mechanism.

\subsection{Evaluation limitations}

The present manuscript validates only the citation-grounding layer
empirically (Section~\ref{sec:eval-cite}), on real arXiv papers. The
result-grounding and revision benchmarks are preliminary and run on
synthetic, illustrative data; the paper does not report a full
applied-science user study, an ablation over individual gates, a
real-data result-grounding evaluation, or an external multi-author
case study. The suite exercises
short drafting and revision tasks on two models from a single
provider; it measures evidence faithfulness (fabricated citations and
numbers, placeholder use, revision drift, gate robustness) rather than
manuscript quality, reviewer calibration, or manuscript-owner
workload, and it enforces the gate through a condensed system prompt
rather than the full released prompt or the orchestration layer below.
Therefore, while the citation benchmark gives direct
evidence that the source-bounded rule suppresses fabrication and
surfaces gaps, the paper does not claim that Paper Pilot improves
overall manuscript quality, decreases review burden, or outperforms
existing writing systems end to end. Its primary contribution remains
a technical and methodological specification for approval-gated
manuscript generation, now accompanied by mechanism-level evidence.

Future work should evaluate Paper Pilot against strong baselines,
including ungated LLM drafting, citation-grounded writing alone, and
conventional human editing workflows. Suggested evaluation tasks
include measuring unsupported-claim leakage, time-to-approval,
manuscript-owner cognitive load, revision traceability, inter-rater
agreement, and reviewer-drift stability. A worked applied-science
case study with a public repository, claim records, evidence
artifacts, and revision logs would provide stronger evidence of
feasibility.

\section{Conclusion}
\label{sec:conclusion}

This paper proposed Paper Pilot, a human-in-the-loop expert system
for evidence-traceable scientific manuscript generation in applied
sciences. The framework addresses a governance problem in
LLM-assisted scientific writing: manuscript claims may propagate from
ideas, methods, evidence, results, and revisions into polished text
without mandatory human approval, traceable support, or explicit
no-pass criteria.

Paper Pilot addresses this problem by adapting the Collaborative
Agent Reasoning Engineering methodology to scientific manuscript
development. In the proposed workflow, LLM agents assist with
section-wise drafting, literature positioning, claim classification,
revision support, and gap identification, while the manuscript owner
remains the accountable authority for approval, override, and final
release. This division of responsibility is central to the framework.
LLM agents are treated as advisory and generative components, not as
final arbiters of scientific validity, methodological adequacy,
evidence sufficiency, or submission readiness.

The framework answers the two research questions through a
structured approval-gated methodology. For RQ1, Paper Pilot defines
mandatory approval gates across the idea-to-claim pipeline, including
scope approval, literature approval, method approval, evidence
approval, claim approval, reviewer override, section approval, and
final release, each with no-pass criteria and audit-log requirements,
specified as a reusable expert-system workflow. For RQ2, the
framework separates LLM reviewer advice from manuscript-owner
authority, reserving scientific validity, methodological adequacy,
evidence sufficiency, and final release as decisions exclusive to the
human author. Two further mechanisms support these answers: an
artifact-level traceability requirement linking manuscript claims to
literature evidence, user-provided findings, or computational
artifacts, and evidence-locked revision control so that revisions
cannot introduce unsupported metrics, unverifiable comparisons, or
unapproved interpretations without renewed review.

The workflow demonstration showed how Paper Pilot can be applied as a
section-wise manuscript-development process. Across drafting
iterations, the system identified missing inputs, produced outlines,
drafted manuscript-ready sections, classified claims, recorded
assumptions, maintained change logs, and surfaced unresolved risks.
This process demonstrated how an LLM-assisted writing workflow can
remain bounded by user-provided evidence and manuscript-owner
decisions rather than relying on autonomous generation alone. As a
first empirical validation, the citation-grounding benchmark---on real
arXiv papers, judge-free---showed that under coverage pressure ungated
drafting fabricated up to a quarter of its citations and never flagged
an evidence gap, while Paper Pilot's evidence-locked rules eliminated
fabricated citations and surfaced the planted gaps as explicit
placeholders. Preliminary benchmarks for result grounding,
evidence-locked revision, and adversarial robustness point the same
way---while also exposing two honest limits, residual revision drift
and a gate breachable under authority escalation by the weaker
model---and their full evaluation on real data and with the complete
released prompt is left to future work.

The contribution of Paper Pilot is therefore methodological and
governance-oriented. It provides a practical expert-system framework
for controlling how LLM-generated manuscript text is produced,
revised, approved, and traced to evidence. For applied-science
researchers, this framework may support more accountable use of LLMs
in manuscript preparation by preserving human authority, reducing
unsupported claim propagation, and maintaining an audit trail across
drafting and revision cycles.

Overall, Paper Pilot positions LLM-assisted scientific writing as a
controlled human-AI decision-support process rather than a fully
autonomous authorship pipeline. By combining CARE-informed agent
engineering, manuscript-owner approval gates, evidence-to-claim
traceability, and evidence-locked revision control, the framework
provides a foundation for accountable, auditable, and scientifically
bounded manuscript generation in applied sciences.

\section*{Declarations}

\subsection*{CRediT authorship contribution statement}

\textbf{Nidhi Jha:} Conceptualization, Methodology, Writing---original
draft, Writing---review \& editing.
\textbf{Siddharth Chaudhary:} Conceptualization, Methodology,
Writing---original draft, Writing---review \& editing.
\textbf{Ajinkya Kulkarni:} Software, Validation, Testing and
benchmarking, Computational implementation and engineering support.

\subsection*{Declaration of generative AI and AI-assisted
technologies in the writing process}

During the preparation of this manuscript, the authors used Paper
Pilot, a human-in-the-loop LLM-assisted scientific writing workflow,
to support section-wise drafting, literature positioning, claim
classification, gap identification, revision tracking, and
manuscript-structure development. This Paper Pilot paper was
implemented using ChatGPT. All AI-generated outputs were reviewed,
revised, and approved by the manuscript owner. The manuscript owner
retained responsibility for the accuracy, integrity, interpretation,
and final approval of all scientific claims.

No AI system was treated as an autonomous author or final authority.
LLM-generated suggestions were used only as drafting and review
support. Claims requiring scientific, methodological, or evidentiary
approval were subject to human review before inclusion in the
manuscript.

\subsection*{Data availability statement}

The data generated for the benchmark suite in
Section~\ref{sec:eval}---scenario definitions, every raw generated
draft across all four benchmarks, and per-run mechanical scoring
records---are released with the paper in the companion repository. No
human-subject or observational dataset was generated or analyzed.

\subsection*{Code and prompt availability}

The Paper Pilot system prompt, approval-gate checklist,
claim-taxonomy template, and audit-log template are available at
\url{https://github.com/nidhi23aug/paperpilot-care}. The benchmark
harness, the programmatic citation fetching and verification tooling
(no LLM writes or verifies bibliography entries), and all scoring
code accompany the paper in the same repository to support
transparency, inspection, and reuse of the proposed workflow.

\subsection*{Funding}

This research received no specific grant from any funding agency in
the public, commercial, or not-for-profit sectors.

\subsection*{Declaration of competing interest}

The authors declare that they have no known competing financial
interests or personal relationships that could have appeared to
influence the work reported in this paper.

\subsection*{Ethics approval and consent to participate}

This study does not report experiments involving human participants,
animals, or identifiable personal data. Ethics approval was not
required because the study did not involve human participants, animal
subjects, or identifiable personal data.

\subsection*{Consent for publication}

Consent for publication was not applicable because the manuscript
does not contain identifiable personal information.

\appendix

\section{Benchmark System Prompts}
\label{app:prompts}

The two system prompts used in the benchmark of
Section~\ref{sec:eval} are reproduced here verbatim. Both conditions
received an identical user prompt (scenario topic, six required
subtopics, the allowed literature set with programmatically fetched
metadata, and the editorial requirement that every subtopic carry at
least one inline citation). The gated prompt operationalizes the
evidence-locked drafting rules of Sections~\ref{sec:workflow}
and~\ref{sec:gates} in condensed form for single-turn drafting; it
is not the full Paper Pilot system prompt.

\subsection*{A.1\quad Ungated condition}

\begin{quote}\small\ttfamily\raggedright
You are an expert academic writer helping draft a journal paper.
Write polished, well-cited scholarly prose. Cite inline in the form
(AuthorSurname et al., YEAR, arXiv:NNNN.NNNNN) and include an arXiv
ID for every citation. You may draw on any literature you know in
addition to the papers provided by the user.
\end{quote}

\subsection*{A.2\quad Gated condition (Paper Pilot drafting rules)}

\begin{quote}\small\ttfamily\raggedright
You are Paper Pilot, a human-in-the-loop scientific writing agent
operating under evidence-locked drafting rules:\\[4pt]
1. SOURCE-BOUNDED LITERATURE: You may cite ONLY the papers
explicitly provided by the user in this conversation. Never cite
any other work, even if you are certain it exists.\\[2pt]
2. NO FABRICATION: Never invent, approximate, or recall-from-memory
a citation, arXiv ID, author list, or finding.\\[2pt]
3. PLACEHOLDER DISCIPLINE: If a requested subtopic is not supported
by the provided papers, do NOT cover it from memory. Instead insert
a placeholder of the exact form [TBD: <what evidence is needed>]
and move on.\\[2pt]
4. CITATION FORMAT: cite inline as (AuthorSurname et al., YEAR,
arXiv:NNNN.NNNNN), using only the metadata supplied by the
user.\\[2pt]
5. NO-PASS AWARENESS: A draft containing a citation outside the
provided set fails its approval gate. Placeholders are acceptable;
unsupported citations are not.
\end{quote}

\bibliographystyle{ieeetr}
\bibliography{references}

\end{document}